\PassOptionsToPackage{table,xcdraw}{xcolor}
\documentclass[manuscript,screen]{acmart}

\AtBeginDocument{%
  }

\setcopyright{acmlicensed}
\copyrightyear{2018}
\acmYear{2018}
\acmDOI{XXXXXXX.XXXXXXX}

\acmConference[Conference acronym 'XX]{Make sure to enter the correct
  conference title from your rights confirmation emai}{June 03--05,
  2018}{Woodstock, NY}
\acmISBN{978-1-4503-XXXX-X/18/06}

\usepackage{booktabs}
\usepackage{graphicx}
\usepackage{adjustbox}
\usepackage{arydshln}
\usepackage{enumitem}
\usepackage{hyperref}
\usepackage{multirow}
\usepackage{makecell} 
\usepackage{tikz}
\usepackage[edges]{forest}

\definecolor{hiddendraw}{RGB}{205, 44, 36}
\definecolor{hidden-blue}{RGB}{194,232,247}
\definecolor{hidden-orange}{RGB}{243,202,120}
\definecolor{hidden-yellow}{RGB}{242,244,193}

\usepackage{amssymb}
\usepackage{pifont}
\newcommand{\cmark}{\ding{51}}%
\newcommand{\xmark}{\ding{55}}%

\begin{document}

\title{A Survey of the Evolution of Language Model-Based Dialogue Systems: Data, Task and Models}

\author{Hongru Wang}
\affiliation{%
  \institution{The Chinese University of Hong Kong}
  \city{Hong Kong SAR}
  \country{China}
  \postcode{999077}}
\email{hrwang@se.cuhk.edu.hk}

\author{Lingzhi Wang}
\affiliation{%
  \institution{The Chinese University of Hong Kong}
  \city{Hong Kong SAR}
  \country{China}}
\email{lzwang@se.cuhk.edu.hk}
  
\author{Yiming Du}
\affiliation{%
  \institution{The Chinese University of Hong Kong}
  \city{Hong Kong SAR}
  \country{China}}
\email{ydu@se.cuhk.edu.hk}

\author{Liang Chen}
\affiliation{%
  \institution{The Chinese University of Hong Kong}
  \city{Hong Kong SAR}
  \country{China}}
\email{lchen@se.cuhk.edu.hk}

\author{Jingyan Zhou}
\affiliation{%
  \institution{The Chinese University of Hong Kong}
  \city{Hong Kong SAR}
  \country{China}}
\email{jyzhou@se.cuhk.edu.hk}

\author{Yufei Wang}
\affiliation{%
  \institution{Huawei Noah Ark Lab}
  \city{Hong Kong}
  \country{China}}
\email{wangyufei44@huawei.com}

\author{Kam-Fai Wong}
\affiliation{%
 \institution{MoE Key Laboratory of High Confidence Software Technologies, The Chinese University of Hong Kong}
 \city{Hong Kong}
 \country{China}}
\email{kfwong@se.cuhk.edu.hk}

\renewcommand{\shortauthors}{Wang et al.}

\begin{abstract}
Dialogue systems (DS), including the task-oriented dialogue system (TOD) and the open-domain dialogue system (ODD), have always been a fundamental task in natural language processing (NLP), allowing various applications in practice. Owing to sophisticated training and well-designed model architecture, language models (LM) are usually adopted as the necessary backbone to build the dialogue system. Consequently, every breakthrough in LM brings about a shift in learning paradigm and research attention within dialogue system, especially the appearance of pre-trained language models (PLMs) and large language models (LLMs). In this paper, we take a deep look at the history of the dialogue system, especially its special relationship with the advancements of language models. Thus, our survey offers a systematic perspective, categorizing different stages in a chronological order aligned with LM breakthroughs, providing a comprehensive review of state-of-the-art research outcomes. What's more, we turn our attention to emerging topics and engage in a discussion on open challenges, providing valuable insights into the future directions for LLM-based dialogue systems. In summary, this survey delves into the dynamic interplay between language models and dialogue systems, unraveling the evolutionary path of this essential relationship. Through this exploration, we pave the way for a deeper comprehension of the field, guiding future developments in LM-based dialogue systems.
\end{abstract}


\thanks{This project is partially supported by HKSAR ITF Project ITT/008/22LP}




\maketitle

\section{Introduction}

Building a conversational intelligent system has always been a fundamental objective in the realm of natural language processing \citep{challenges_in_open_domain,challenges_in_tods}. The ability to interact naturally and seamlessly with machines has opened up new avenues for human-machine communication, leading to transformative applications across different industries, from \texttt{Siri}, \texttt{Xiaoice} \citep{zhou2020design} to the latest \texttt{New Bing} \footnote{\url{https://www.bing.com/new}} and \texttt{Google Bard} \footnote{\url{https://bard.google.com/}}. At the heart of this evolution is the pivotal role that dialogue systems play, which are designed to facilitate interactions characterized by harmlessness \citep{instructgpt}, helpfulness \citep{cuecot}, trustworthiness \citep{Huang_hallucination_survey}, and personalization \citep{salemi2023lamp}. These systems aim to emulate human-to-human conversations, thereby offering the potential to enhance user experiences, streamline tasks, and provide personalized assistance across domains such as customer support, virtual assistants, healthcare, education, and more, depending on the types of dialogue context \cite{types_of_dialogs}. Throughout the entire evolution of conversational intelligent systems, two factors have been pivotal: \textit{data} and \textit{model}.

\textit{Data.} Specifically, there are two major types of dialogue data in practice: \textit{task-oriented dialogue} and \textit{chit-chat}. Task-oriented dialogue revolves around assisting users in achieving specific tasks or goals, such as making reservations or booking tickets. In this context, the dialogue system acts more like a helpful assistant, providing relevant information and guidance to users \citep{SimpleToD,challenges_in_tods}. Conversely, chit-chat involves casual and informal conversations, mainly aimed at establishing and maintaining social connections. Here, the dialogue system adopts the role of a friendly chatbot, engaging users in a more relaxed and conversational manner \citep{challenges_in_open_domain}. Based on their distinct roles, two types of dialogue systems have been proposed: \textbf{Task-oriented Dialogue System (TOD)} and \textbf{Open-domain Dialogue System (ODD)}. The TOD is designed to efficiently handle task-oriented conversations, guiding users towards accomplishing specific objectives by detecting user intentions (natural language understanding, NLU), tracking dialogue state (dialogue state tracking, DST), making suitable actions (dialogue policy learning, DPL), and responding accordingly (natural language generation, NLG). In contrast, the ODD is intended for open-domain interactions, enabling free-flowing conversations on a wide range of topics by directly mapping the dialogue context into the response, without a predefined task or goal. These two types cater to different use cases and user needs, showcasing the versatility and applicability of dialogue systems in diverse scenarios. Thus, most of the previous works (\textit{before $\sim$2020}) aim to design and build these dialogue systems independently using different backbones considering the intrinsic characteristics of different types of data~\cite{challenges_in_tods, challenges_in_open_domain}. 

\textit{Model.} The revolutionary advancements in language models (LMs) have catalyzed a transformative evolution in numerous natural language tasks, including dialogue systems, reshaping their very foundation. The core objective of LMs is to predict word sequence probabilities has brought about a profound impact on dialogue systems~\citep{tran-etal-2017-neural,gpt}, empowering them to learn world knowledge from the pre-training corpus \citep{petroni2019language} and generate more contextually relevant and helpful responses \citep{zhang-etal-2020-dialogpt}. Concurrently, each milestone in LM development brings paradigm shift in dialogue systems, playing a pivotal role in their rapid evolution and ongoing trends, especially the appearance of latest large language models (LLMs), such as ChatGPT. 

\begin{figure*}[t]
    \centering
    \includegraphics[trim={0cm 4.5cm 0cm 3.5cm}, clip, scale=1.0, width=0.9\textwidth]{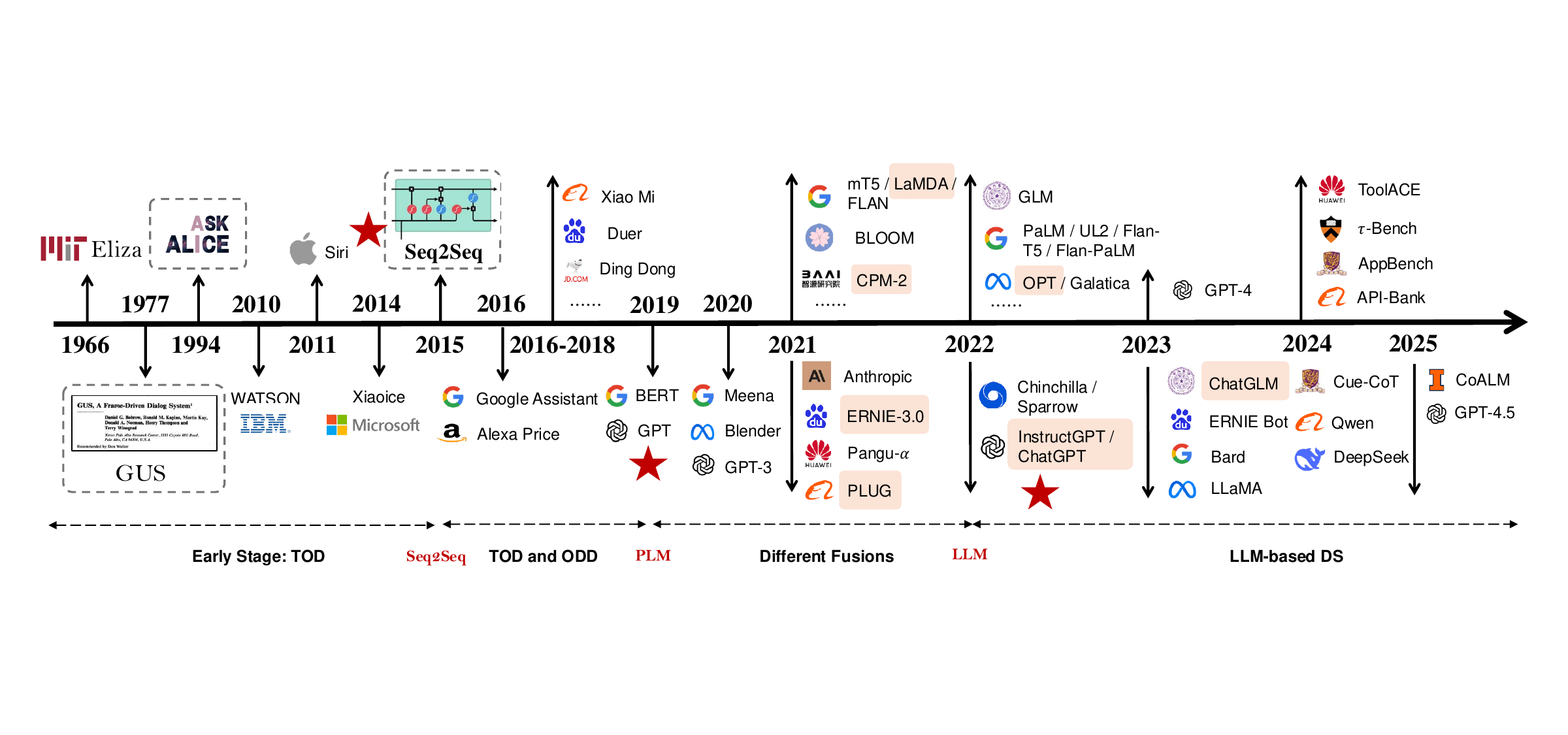}
    \caption{The development of LM-based Dialogue Systems, which can be divided into four distinct stages, each marked by significant breakthroughs that have paved the way for advancements in the field: 1) Early Stage (1966 - 2015); 2) The Independent Development of TOD and ODD (2015 - 2019); 3) Fusions of Dialogue Systems (2019 - 2022); 4) LLM-based DS (2022 - Now).}
    \label{fig:evolution_ds}
\end{figure*}

Rather than focusing solely on the development of dialogue systems from a data perspective or the advancement of language models from a model perspective, as previous surveys have done \citep{survey_ds, systematic_survey}, we emphasize the dynamic interaction between data and models. This survey highlights the fascinating evolution of dialogue systems through this transformative interplay, showcasing a clear path from traditional rule-based dialogue system towards current dominated LLM-based dialogue systems. This whole progression can be categorized into four major stages, corresponding to the four developmental phases of language models as illustrated in Figure~\ref{fig:evolution_ds}.

\textit{Early stage -- Statistical language models (SLMs).} The first dialogue system --- Eliza, was proposed by MIT in 1966, which was even earlier than the rise of SLMs in the 1990s, allowing plausible conversations between humans and machines. Subsequently, a series of dialogue systems were constructed, including a travel agent aiding the client to make a simple return trip to a single city in California \cite{bobrow1977gus} and a rule-based chatbot configured with a persona, employing heuristic pattern matching rules to address inquiries about ages, hobbies, interests, and etc \cite{alice} \footnote{\url{https://web.njit.edu/~ronkowit/eliza.html}}. In addition to academic endeavors, more and more companies pay attention to the field of dialogue systems and have developed their own products. For example, IBM has developed \texttt{WATSON}, Apple has \texttt{Siri}, and Microsoft has \texttt{XiaoIce}. These early virtual assistants have limited functionalities and follow a rigid process, primarily assisting users with single and simple tasks like scheduling appointments, setting reminders, and providing basic answers to questions. At this stage, most of these dialogue systems are task-oriented, the prevailing nature of these dialogue systems is predominantly task-oriented, employing either modular-based approaches or machine-learning methods based on SLMs.

\textit{The independent development of TOD and ODD} --- \textit{Neural language models (NLMs).} Around 2015, there was a sea change when the sequence to sequence (seq2seq) framework was proposed, in which salient features were learned jointly with the training of the model --- NLM \cite{bengio2000neural,mikolov2010recurrent}. Specifically, NLMs characterize the probability of word sequences by neural networks, e.g., two typical recurrent neural networks (RNNs) such as long short-term memory networks (LSTM) \citep{lstm} and gated recurrent units (GRU) \citep{gru}. These NLMs serve as the backbone to map variable-length input sequences to variable-length output sequences which is achieved by employing two main components: an encoder and a decoder. This paradigm shift facilitated the emergence of Open-domain Dialogue (ODD) systems ($\sim$2015), garnering attention due to a well-defined end-to-end framework enabled by the seq2seq network. Simultaneously, NLMs were also applied to TOD, with the goal of achieving better performance and performing complex tasks in multiple domains \citep{peng2017composite,wen-etal-2017-network}.

\textit{Fusions of Dialogue Systems} ---  \textit{pre-trained language models (PLMs).} Traditional NLMs requires extensive hand-crafted features selection and domain-specific knowledge, making them cumbersome and time-consuming to develop for each task. However, with the advent of deep learning and large-scale language corpora, \textbf{pre-trained language models (PLMs)} emerged as a groundbreaking solution. These models are pre-trained on massive amounts of unlabeled text data, capturing rich semantic and syntactic patterns \citep{devlin-etal-2019-bert,gpt,gpt2}. By leveraging pre-training, subsequent fine-tuning on specific tasks becomes more accessible, enabling the models to adapt quickly and achieve impressive results with minimal task-specific data. In the realm of dialogue systems, DialoGPT exemplifies this approach by undergoing pre-training on 147 million conversation-like exchanges extracted from Reddit comment chains and then finetuning on a few training examples to generate more relevant, informative, and context-consistent responses \cite{zhang-etal-2020-dialogpt}. This paradigm, commonly known as \textit{pre-train and finetune} \cite{pengfeiliu_survey}, leads to the resulting model as \textbf{pre-trained dialogue model (PDM)} that undergoes fine-tuning using a dialogue corpus. It is worth noting that the major differences between PDMs lies in the the data used for training and the backbone language models adopted, such as PLATO \citep{plato,plato-2}, BlenderBot \citep{blenderbot}, Meena \citep{meena}, Pangu-bot \citep{mi2022pangubot}, PLUG \citep{bi-etal-2020-palm}, and etc. During this stage, there is an increasing focus on the different level of fusion: 1) different sub-tasks inside TOD, resulting in end-2-end TOD \citep{qin-etal-2023-end}, 2) TOD with ODD, resulting in unified dialogue system.


\textit{LLM-based dialogue system} --- \textit{Large language models (LLM).} Recognizing the correlation between increased pre-training corpus size and model size with enhanced performance across various NLP tasks \citep{kaplan2020scaling}, researchers try to scale the model size and the size of pre-training corpus simultaneously to be more sample efficient, enabling the model to learn more intricate patterns and representations from the data. Thus, the PLMs become \textbf{Large Language Models (LLMs)}, such as GLM \citep{du2022glm}, LLaMA \cite{touvron2023llama}, and InstructGPT \citep{instructgpt}. Due to the larger model size and the utilization of extensice, high-quality pre-training corpora, these LLMs offer unprecedented capabilities in many language understanding and generation tasks, including but not limited to question-answering and named entity recognition, thus reshaping the landscape of dialogue systems. With further instruction-tuning using data as \texttt{(instruction, input, output)}, these LLMs are capable of answering diverse questions and following various instructions such as writing an email and telling a joke. At this moment, the LLMs can be directly used as the dialogue systems. Moreover, finetuning the LLM with in-domain data, i.e., dialogue/conversational corpus, can further enhance its conversational ability. For example, the evolution of GPT-3.5 to InstructGPT, and then ChatGPT exactly follows this path. Similarly, ChatGLM \citep{zhang2023glm} is also finetuned using publicly available Chinese dialogue datasets based on the corresponding LLM -- GLM. To enhance alignment with human preferences and values, the incorporation of reinforcement learning becomes imperative, particularly in the final stages of model refinement. This involves leveraging feedback from either human evaluators \citep{instructgpt} or automated systems \citep{mine_icassp} to iteratively improve the model's performance and ensure its adherence to desired criteria. After all, this type of \textbf{LLM-based dialogue system (LLM-based DS)} performs well at both task-oriented dialogue and chit chat, making it an ideal foundation for a general conversational AI.
    
It is worth noting that the development of \textbf{LM-based Dialogue System} is a ongoing and evolving process, and the stages are not rigidly separated by specific turning points. Instead, there are overlaps and continuous advancements across these stages, particularly in the later stages. Generally, with the advancement of LM, the boundary between TOD and ODD and the boundary between dialogue model and language model have become increasingly blurred, ushering in a new era of LLM-based dialogue system. Despite that the existing studies have thoroughly examined the development and challenges associated with different types of dialogue systems, such as TOD \cite{challenges_in_tods} and ODD \cite{challenges_in_open_domain}, there still remains a notable gap in the literature when it comes to offering a holistic perspective on the evolutionary trajectory of dialogue systems, particularly in light of the advancements in LM. By critically analyzing and understanding the impact of pivotal milestones, we can gain valuable insights into the possibilities that await and the transformative impact dialogue systems may have on diverse domains and applications in the foreseeable future. This survey aims to present a comprehensive overview of the realm of LM-based dialogue systems \footnote{We have made thorough efforts to cover a wide range of works, but acknowledge the possibility of omissions. Readers are encouraged to provide suggestions for any missed contributions or errors. We are committed to updating the article with new approaches or definitions proposed by the community.}, shedding light on the prominent directions which the field is currently progressing. This entails a meticulous review of the existing literature, aimed at elucidating the paradigm shifts observed within different stages of LM-based dialogue systems. These shifts encompass a range of integrations across various levels, notably encompassing the data, model and task dimensions. In order to highlight this evolution path, we will not focus on specific minor modifications of model for different tasks (i.e., LSTM and GRU) but instead showcase how these fusion happens with advancement of language models~\citep{tay2022scalinglawsvsmodel}. We emphasize this is more important at the era of large language models, since most of previous modifications at the large language model level becomes inefficient and infeasible, resulting in unpredictable consequence.



The remainder of this survey is organized as follows: Section~\ref{background} provides the description of different types of dialogues, unified task definition and corresponding commonly used metrics and benchmarks in dialogue system, followed by the early development of two types of LM-based dialogue systems: TOD and ODD in Section~\ref{independent_development}. Sections ~\ref{fusion_of_ds} review and summarize the evolution of different dialogue systems when PLMs are scaled to LLMs. We then investigate emerging trends and open challenges in LLM-based DS (See Section~\ref{llm-based_ds}). Then, Section~\ref{discussion} discusses the open problems for LLM-based dialogue systems. Finally, we conclude the survey in Section~\ref{conclusion} by summarizing the major findings and discussing the remaining issues for future work.

    

\begin{table*}[t]
    \centering
    \caption{Different types of dialogues in TOD and ODD. We keep the last turn of users for conciseness and brevity.}
    \label{tab:types_dialogs}
    \begin{adjustbox}{max width=0.8\textwidth}
    \begin{tabular}{c|c|p{0.2\linewidth}|p{0.3\linewidth}|c}
    \toprule
    \textbf{Types of DS} & \textbf{Types of Dialogues} & \textbf{User message} (\textit{U}) & \textbf{System response} (\textit{S})  & \textbf{External knowledge} (\textit{K})  \\
    \hline
    
    TOD & Task-oriented & I need to find a nice restaurant in Madrid that serves expensive Thai food & There is a restaurant called Bangkok City located at 9 Red Ave. & Restaurant database \\
    
    \hline

    & Social Chit-Chat & How are you going ? & I am good. And you? & - \\
    \cdashline{2-5}
    \multirow{3}{*}{ODD} & KG Chit-Chat &  I like watching action movies & Here are some action movies from various eras and styles that you might enjoy: \textit{Inception}, \textit{The Avengers} & Movie KG \\
    \cdashline{2-5}
    & Question Answering & Do you know which team wins the 2022 World Cup? & The Argentina team & Wikipedia \\
    
    \bottomrule
    \end{tabular}
    \end{adjustbox}
    \vspace{-5mm}
\end{table*}


\section{Background}
\label{background}

Before we proceed, it is essential to figure out how many different types of dialogs are in our daily life. Table~\ref{tab:types_dialogs} demonstrates some examples of four major types of dialogs, including:

\begin{enumerate}[label=(\alph*)]

\item \textbf{social chit-chat.} This type of dialog involves casual and informal conversations aimed at establishing and maintaining social connections. It includes light-hearted discussions about topics such as current events \citep{emh,psyqa}, personal experiences \citep{personachat,dulemon}, hobbies, and interests \citep{weibo}. Social chit-chat is typically used to create a friendly atmosphere and build social bonds, requiring the DS to generate persona-consistent, empathetic, and compassionate responses besides helpful.

\item \textbf{knowledge-grounded chit-chat.} Knowledge-grounded dialogs revolve around exchanging information and discussing topics that require a certain level of expertise or specific knowledge. It can involve discussing academic subjects, professional domains, or specialized areas of interest. This type of dialog often focuses on sharing and acquiring knowledge \citep{wow,zhou-etal-2020-kdconv}. The knowledge can be formed in different formats such as graphs \citep{comet}, tables \citep{table_qa}, figures \cite{vqa}, or even a combination of them \citep{more_is_better}. Most of previous methods for this type of dialog based on retrieval-augmented generation (RAG) framework \citep{rag}.

\item \textbf{question answering.} Question-answering dialogs involve a user asking specific questions, and the dialogue system providing relevant and accurate answers. These dialogs are typically focused on obtaining specific information or clarifications. The system's role is to understand the user's queries and provide appropriate responses based on the available knowledge or resources \citep{psyqa,lin2022truthfulqa}.

\item \textbf{task-oriented dialogue.} Task-oriented dialogues center around assisting users in accomplishing specific tasks or goals. Users seek help from the dialogue system to perform actions such as making reservations, booking flights, ordering products, or seeking recommendations. These dialogues often involve a series of steps or sub-tasks that need to be completed to achieve the overall objective \citep{peng2017composite, budzianowski2018multiwoz, zhu2020crosswoz}.
\end{enumerate}

\vspace{-2mm}
These are general descriptions of the different types of dialogue, and there may be variations or combinations of these types in real-world dialogue scenarios \citep{types_of_dialogs}. The diversity and complexity of each type of dialogue not only contribute to the wide range of applications for dialogue systems but also give rise to various challenges and problems. These challenges pose unique requirements and considerations, necessitating the development of dialogue systems that are both versatile and robust that can effectively handle the intricacies of each dialogue scenario in a unified manner. 

\subsection{Task Definition}
\label{task_definition}

Following the unified definition recently \cite{types_of_dialogs,zhang2023dialogstudio}, we formulate the task of dialogue system as a seq2seq problem by directly mapping the dialogue context to the final response with or without the assistance of different external sources such as the local database in TOD, different knowledge bases in ODD, \textit{e.g.,} persona and documents. Formally, given the dialogue context $c = \{u_1, s_1, ..., u_t\}$, and optionally the external knowledge bases $K$, the goal of the dialogue system is to generate a helpful and harmless response $r$ as shown below:

\vspace{-4mm}
\begin{equation}
    r = \boldsymbol{M} (c, K) \quad or \quad \boldsymbol{M} (c)
\end{equation}

Where $\boldsymbol{M}$ is parameterized by different language models. Specifically, there are significant differences in interactions between various dialogue systems with the external knowledge base $K$. In this work, we discuss two major interactions corresponding to TOD and ODD respectively. Formally, the TOD first generates the belief state according to the given context \citep{zhao2021unids, peng-etal-2021-soloist}:

\vspace{-5mm}
\begin{equation}
    b = \boldsymbol{M} (c)
\end{equation}

And then use it as a query to search related information in the local database $K$. The returned result $s$ is used as additional information to generate the final response, the entire process can be defined as follows:

\vspace{-4mm}
\begin{equation}
    TOD (r | c, K) = f (r | c, b, s) f (b | c)
\end{equation}

Secondly, for the ODD, the processing is relatively simpler by directly rewriting the dialogue context as the query or using the whole dialogue context to search the external knowledge base $K$ \citep{wow}. In this case, TOD can be regarded as one special type of ODD with the source of knowledge as corresponding database \citep{q-tod}.

\vspace{-5mm}
\begin{equation}
    ODD (r | c, K) = f (r | c, q, s) f (q | c)
\end{equation}

It is important to clarify here that ODD contains the cases which does not require any external knowledge, by directly mapping dialogue context to corresponding responses using LMs \citep{survey_ds}.

\subsection{Metrics and Benchmarks}

To evaluate the performance of different types of dialogue systems, several metrics and benchmarks are carefully designed. In order to assess various aspects of dialogue generation, including task success, response quality, and consistency. Depending on the type of dialogue system, the evaluation metrics and benchmarks may differ significantly. Table~\ref{tab:metrics_ds} shows the common metrics and benchmarks used in the dialogue systems.

\begin{table*}[!t]
    \renewcommand{\arraystretch}{1.0}
    \caption{Metrics and benchmarks for different tasks in dialogue systems.}
    \label{tab:metrics_ds}
    \centering
    \begin{adjustbox}{max width=0.8\textwidth}
    \begin{tabular}{clccc}
    \toprule
    \textbf{DS} & \textbf{Task} & \textbf{Benchmarks} & \textbf{Metrics} & \textbf{Evaluation Type} \\
    \hline
    
    \multirow{6}{*}{TOD} & 
    NLU & SNIPS~\citep{coucke2018snipsvoiceplatformembedded}, ATIS, Few-NERD~\citep{ding-etal-2021-nerd} & Acc, F1 & Automatic \\

    \cline{2-5}
    & DST & DSTC-\{2,4\}~\citep{henderson-etal-2014-second, kim2017fourth}, Wizard-of-Oz~\citep{mrksic-etal-2017-neural}, CoSQL~\citep{yu-etal-2019-cosql}  & Joint-Acc & Automatic \\

    \cline{2-5}
    & DPL & MultiWoZ~\citep{eric-etal-2020-multiwoz}, RisaWoZ~\citep{quan-etal-2020-risawoz}, CrossWoZ~\citep{zhu2020crosswoz} & Acc, F1, Reward & Automatic \\

    \cline{2-5}
    & NLG & E2E~\citep{novikova-etal-2017-e2e}, BAGEL~\citep{mairesse-etal-2010-phrase}, RNNLG~\citep{wenmultinlg16}, FewshotWoZ~\cite{peng-etal-2020-shot} & BLEU, Rouge, METEOR & Automatic \\
    
    \cline{2-5}
    & \multirow{3}{*}{General} & MultiWoZ~\citep{eric-etal-2020-multiwoz}, RisaWoZ~\citep{quan-etal-2020-risawoz}, CrossWoZ~\citep{zhu2020crosswoz}, SGD~\citep{rastogi2020towards}  & \multirow{3}{*}{Succ, Turns} & \multirow{3}{*}{Automatic} \\
    & & KddRES~\citep{wang2021kddres}, CamRest676~\citep{wen2017latent}, Frames~\citep{el-asri-etal-2017-frames}, KvRET~\citep{eric2017key} &  &  \\
    & & \textit{Multilingual:} BiToD~\citep{lin2021bitod}, X-RisaWoZ~\citep{moradshahi-etal-2023-x}, IndoToD~\citep{kautsar-etal-2023-indotod} & \\

    \hline
    \multirow{8}{*}{ODD} & Retrieval & WoW~\citep{wow}, FoCus~\citep{focus}, KdConv~\citep{zhou-etal-2020-kdconv}, KBP~\citep{wang2023large}  & Recall@k, Hit@k & Automatic \\

    \cline{2-5}
    & \multirow{7}{*}{NLG} & \textit{Chit-chat:} DailyDialog~\citep{li-etal-2017-dailydialog}, Topical-Chat~\citep{gopalakrishnan2023topical}, CMU\_DoG~\citep{zhou2018dataset} &  &  \\
    & & \textit{Empathy:} EmpatheticDialogues~\citep{ed}, ESConv~\citep{liu2021towards}, D4~\citep{d4} & \\
    & & \textit{Persona:} PersonaChat~\citep{personachat}, KvPI~\citep{song-etal-2020-profile}, PersonalDialog~\citep{zheng2019personalized}, FoCus~\citep{focus}  & BLEU, Rouge, PPL, Distinct & Automatic \\
    & & \textit{Memory:} DuLeMon~\citep{dulemon}, MSC~\citep{xu-etal-2022-beyond}, PerLTQA~\citep{du-etal-2024-perltqa} & Fluency, Coherence & Human Evaluation  \\
    & & \textit{Knowledge:} WoW~\citep{wow}, KdConv~\citep{zhou-etal-2020-kdconv}, OpenDialKG~\citep{moon-etal-2019-opendialkg}, KBP~\citep{wang2023large} & Win Rate & LLM-as-a-Judge  \\
    & & \textit{Multilingual:} XPersona~\citep{lin-etal-2021-xpersona}, XDailyDialog~\citep{liu-etal-2023-xdailydialog}  & \\
    & & \textit{Multimodal:} Image-Chat~\citep{shuster-etal-2020-image}, MMChat~\citep{zheng-etal-2022-mmchat}, StickerConv~\citep{zhang-etal-2024-stickerconv} & \\

    \cline{2-5}
    & \multirow{2}{*}{NLI} & PersonaChat~\citep{personachat}, KBP~\citep{wang2023large}, KvPI~\citep{song-etal-2020-profile} & \multirow{2}{*}{Consistency} & Automatic \\
    &  & Dialogue NLI~\citep{welleck-etal-2019-dialogue}, FoCus~\citep{focus}  &  & LLM-as-a-Judge \\

    \bottomrule
    \end{tabular}
    \end{adjustbox}
    \vspace{-6mm}
\end{table*}

\subsubsection{Task-oriented Dialogue System.}

A modular task-oriented dialogue system consists of several sub-tasks, each responsible for a specific function. Similarly, an end-to-end task-oriented dialogue system performs these functions implicitly, but the evaluation metrics used to assess each module in a modular system can also be applied to evaluate the corresponding aspects of an end-to-end system.

\paragraph{Natural Language Understanding (NLU)} NLU processes user utterances and extracts relevant semantic frames, consisting of two subtasks: intent classification and slot filling. NLU can be rule-based, where predefined patterns or rules are used to extract information, or it can employ deep learning techniques like intent classification and named entity recognition. The result usually includes the intent of user and corresponding slots information such as departure location and destination if the user's intent is detected as book flight ticket. 

\begin{itemize}
    \item Metrics. To evaluate the performance for these two tasks (i.e., intent detection and slot filling), Accuracy and F1-score are widely employed as defined as follows:

    \vspace{-2mm}
    \begin{equation}
        Acc = \frac{\sum_{i=1}^{N} \mathbf{1}(\hat{y}_i = y_i)}{N} \quad 
        \text{F1} = 2 \times \frac{\text{Precision} \times \text{Recall}}{\text{Precision} + \text{Recall}}
    \end{equation}

    where $\hat{y}_i$ is the predicted label for the $i_{th}$ instance, $y_i$ is the ground truth label, $N$ is the total number of instances and $\mathbf{1}$ is the indicator function.

    \item Benchmarks. SNIPS~\citep{coucke2018snipsvoiceplatformembedded} and ATIS~\citep{atis} are two classic and commonly used datasets for intent detection and slot filling tasks in NLU, and there are some other datasets such as Few-NERD~\citep{ding-etal-2021-nerd} which focus on few-shot named entity recognition.
\end{itemize}

\paragraph{Dialogue State Tracking (DST)} DST maintains and updates the state of the conversation as it progresses. It keeps track of the user's goals, preferences, and any relevant context obtained from the dialogue context by updating the intent and slots information. DST can use a rule-based approach to update according to the results from the NLU module, or replace the NLU module and directly track the dialogue state from a dialogue context with similar architecture as NLU \citep{balaraman-etal-2021-recent, wang-etal-2021-fast}. 

\begin{itemize}
    \item Metrics. It is crucial to track the dialogue states in terms of joint accuracy of the intent and slots as the conversation progresses. Joint accuracy measures the correctness of the dialogue state at each turn, considering it correct only if all intents and slot values are predicted accurately~\citep{henderson-etal-2014-second}.

    \item Benchmarks. The Dialogue System Technology Challenges (DSTC) have introduced several well-known benchmarks like DSTC2~\citep{henderson-etal-2014-second} and DSTC4~\citep{kim2017fourth}, primarily focusing on dialogue state tracking. In addition to these, there are more advanced benchmarks designed to tackle complex scenarios, including multi-domain dialogue state tracking (Wizard-of-Oz~\citep{mrksic-etal-2017-neural}) and tasks grounded in SQL for more sophisticated state management (CoSQL~\citep{yu-etal-2019-cosql}).
\end{itemize}

\paragraph{Dialogue Policy Learning (DPL)} The dialogue policy component determines the system's actions or responses based on the current dialogue state. It maps the dialogue state to appropriate system actions, which can include asking clarification questions, providing recommendations, or executing relevant API calls. Dialogue policies can be rule-based, handcrafted, or learned using techniques like reinforcement learning \citep{peng2017composite,gdpl,takanobu-etal-2020-multi,wang-etal-2020-learning-efficient,mine_icassp} or supervised learning \citep{henderson_hybrid_2008,weibo}. More details can be found in the survey paper \citep{mine_mir}.

\begin{itemize}
    \item Metrics. Besides the above mentioned Accuracy and F1, DPL additionally uses reward score as one of evaluation metrics to evaluate whether or not current policy leads to greater reward scores.

    \item Benchmarks. It is common to directly use some general task-oriented dialogue corpus to evaluate the performance of dialogue policy, including but not limited to MultiWoZ~\citep{eric-etal-2020-multiwoz}, RisaWoZ~\citep{quan-etal-2020-risawoz} and CrossWoZ~\citep{zhu2020crosswoz}.
\end{itemize}

\paragraph{Natural Language Generation (NLG)} NLG generates system responses in natural language based on the dialogue policy's output. It transforms structured information or system actions into coherent and fluent sentences that can be understood by users. NLG can employ template-based approaches, rule-based generation, or more advanced techniques like neural network-based generation models \citep{slot-consistent_nlg,zhu-etal-2020-convlab}.

\begin{itemize}
    \item Metrics. Ultimately, these generated response can be directly compared with the ground truth response in the original dataset. The quality is commonly measured using BLEU~\citep{papineni-etal-2002-bleu}, ROUGE~\citep{lin-2004-rouge}, and sometimes METEOR~\citep{banerjee-lavie-2005-meteor}.

    \item Benchmarks. The commonly used benchmarks for natural language generation (NLG) in task-oriented dialogue systems include E2E NLG~\citep{novikova-etal-2017-e2e}, BAGEL~\citep{mairesse-etal-2010-phrase}, RNNLG~\citep{wenmultinlg16}, FewshotWoZ~\cite{peng-etal-2020-shot}. Among these, E2E NLG and BAGEL focus on single-domain scenarios, while RNNLG and FewshotWoZ are designed for multi-domain settings.
\end{itemize}

\paragraph{General} In addition to independently evaluating the performance of each sub-task in a modular task-oriented dialogue system, there are several general metrics and benchmarks available for assessing the overall performance of the entire system. To ensure a comprehensive and consistent evaluation, ConvLab \citep{zhu-etal-2020-convlab} has been introduced as an open-source toolkit. It provides researchers with the capability to build task-oriented dialogue systems and evaluate the performance of each component effectively.

\begin{itemize}
    \item Metrics. The primary objective of a task-oriented dialogue system is to assist users in completing predefined tasks, making the \textit{success rate} the most critical metric, defined as the proportion of dialogues in which the system successfully helps users achieve their goals. Another important metric is the \textit{average number of dialogue turns}, as an effective system should complete tasks in as few turns as possible, minimizing unnecessary interactions.

    \item Benchmarks. There are different types of task-oriented dialogue corpus in different applications: i) general domain: MultiWoZ~\citep{eric-etal-2020-multiwoz}, RisaWoZ~\citep{quan-etal-2020-risawoz}, CrossWoZ~\citep{zhu2020crosswoz}; and ii) specific domain: KddRES~\citep{wang2021kddres}, CamRest676~\citep{wen2017latent}, Frames~\citep{el-asri-etal-2017-frames}, KvRET~\citep{eric2017key}; and iii) multilingual: BiToD~\citep{lin2021bitod}, X-RisaWoZ~\citep{moradshahi-etal-2023-x}, IndoToD~\citep{kautsar-etal-2023-indotod}. Specifically, KddRES~\citep{wang2021kddres} is the first Cantonese task-oriented dialogue corpus on restaurant, while BiToD~\citep{lin2021bitod} is the first bilingual multi-domain dataset for end-to-end task-oriented dialogue modeling. All of these benchmarks can be used for evaluation of specific sub-tasks or general end-to-end performance.
\end{itemize}

\subsubsection{Open-domain Dialogue System} Open-domain dialogue systems (ODDs) are designed to engage in conversations on a wide range of topics without domain-specific constraints. Evaluating the performance of these systems is challenging due to the complexity and variability of dialogue responses. To ensure comprehensive assessment, evaluation typically involves three key tasks: Knowledge Retrieval, Natural Language Generation (NLG), and Natural Language Inference (NLI). Each task has dedicated metrics and benchmarks that measure system performance from different perspectives. While different types of ODDs may only involve one or two of these sub-tasks, we showcase the common process typically integrates all three to ensure accurate and comprehensive evaluation.

\paragraph{Knowledge Retrieval.} Knowledge retrieval is a fundamental step in ODDs, particularly for knowledge-grounded dialogue systems that rely on external sources to generate informed and contextually relevant responses. The goal is to retrieve the most relevant knowledge from a large-scale knowledge base or document collection.

\begin{itemize}
    \item Metrics. Retrieval tasks typically use \textit{Recall@k} and \textit{Hit@k}, measuring how often the correct candidate response appears in the top-k retrieved items. For example, Recall@k is:

    \[
    \text{Recall@k} = \frac{\text{Number of times the correct candidate is in top } k}{\text{Total number of queries}},
    \]

    \item Benchmarks. Many knowledge-augmented dialogue corpora include additional annotations to support retrieval tasks. Examples include WoW \citep{wow}, FoCus \citep{focus}, KdConv \citep{zhou-etal-2020-kdconv}, and KBP \citep{wang2023large}. These datasets often provide relevant documents, sentences, or knowledge graphs to facilitate effective information retrieval and dialogue generation.
\end{itemize}

\paragraph{Natural Language Generation (NLG)} After optionally retrieving relevant knowledge, the system can generate a response. Since open-ended dialogues allow for multiple valid responses, it is unreasonable to only consider single ground truth answer. This makes evaluating response quality challenging due to the diversity of plausible responses.

\begin{itemize}
    \item Metrics. There are three different ways to evaluate the generated response for open-domain dialogue system. First of all, there are several automatic evaluation metrics, including \textit{BLEU, Rouge, PPL and Distinct-N}. Secondly, since the content of dialogue is open-ended, there is no strict ground truth response given the dialogue context. Thus, it is usually required to conduct additional \textit{human evaluation} to directly employ human annotators to grade the quality of generated responses based on different aspects, such as fluency and coherence. Recently, as LLMs demonstrate a high level of agreement with human judgments, they are increasingly being used to replace human annotators in grading responses (i.e., \textit{LLM-as-a-judge}). For example, these LLMs can be employed to directly compare responses generated by different systems, ultimately determining the winning model based on win rate.

    \item Benchmarks. There are numerous benchmarks designed to evaluate the performance of open-domain dialogue systems. These benchmarks can be categorized based on whether they rely on external knowledge and the specific types of knowledge sources they use. Some datasets, such as those for chit-chat (DailyDialog~\citep{li-etal-2017-dailydialog}, Topical-Chat~\citep{gopalakrishnan2023topical},) and empathetic dialogues (EmpatheticDialogues~\citep{ed}, ESConv~\citep{liu2021towards}, D4~\citep{d4}), do not require external knowledge, focusing instead on generating engaging and emotionally aware conversations. In contrast, other datasets are designed to incorporate various types of external knowledge into responses, including world knowledge (WoW~\citep{wow}, KdConv~\citep{zhou-etal-2020-kdconv}, KBP~\citep{wang2023large}), persona (PersonaChat~\citep{personachat}, KvPI~\citep{song-etal-2020-profile}, PersonalDialog~\citep{zheng2019personalized}), and memory (DuLeMon~\citep{dulemon}, MSC~\citep{xu-etal-2022-beyond}, PerLTQA~\citep{du-etal-2024-perltqa}). Additionally, several benchmarks explore more complex scenarios, such as multilingual conversations (XPersona~\citep{lin-etal-2021-xpersona} and XDailyDialog~\citep{liu-etal-2023-xdailydialog}) and multi-modal interactions (MMChat~\citep{zheng-etal-2022-mmchat}).
\end{itemize}

\paragraph{Natural Language Inference (NLI)} Sometimes, it is required to assess whether the generated response is consistent with the retrieved knowledge. A dedicated NLI model is often used to classify the relationship between the response and the knowledge as consistent, contradictory, or neutral.
 
\begin{itemize}
    \item Metrics. The consistency score can be calculated as the proportion of responses classified as consistent out of all generated responses. Depending on specific type of incorporated knowledge, there are different types of consistency scores such as persona consistency, knowledge consistency and memory consistency.

    \item Benchmarks. It is worthy noting that mostly the NLI benchmark can be extracted from existing open-domain dialogue corpus if it contains the augmentation of external knowledge, such as KBP~\citep{wang2023large} and FoCus~\citep{focus}. In addition, there are several general dialogue NLI datasets such as Dialogue NLI~\citep{welleck-etal-2019-dialogue}.
\end{itemize}


\section{The Independent Development of TOD \& ODD}
\label{independent_development}


\begin{figure*}[tp]
  \centering
  \begin{forest}
    forked edges,
    for tree={
      grow=east,
      reversed=true,
      anchor=base west,
      parent anchor=east,
      child anchor=west,
      base=left,
      font=\small,
      rectangle,
      draw=hiddendraw,
      align=left,
      minimum width=2.5em,
      s sep=6pt,
      inner xsep=2pt,
      inner ysep=1pt,
      ver/.style={rotate=90, child anchor=north, parent anchor=south, anchor=center},
    },
    where level=1{text width=2em,font=\scriptsize}{},
    where level=2{text width=6em,font=\scriptsize}{},
    where level=3{text width=14em,font=\scriptsize}{},
    where level=4{text width=12em,font=\scriptsize}{},
    [LM-based\\Dialogue System
    [TOD
        [Natural Language\\Understanding (NLU)
           [RNN: RNN-LU \citep{yao2013recurrent, mesnil2014using} \\
           LSTM: BiLSTM \citep{yao2014spoken, hakkani2016multi}{,} Attention-BiLSTM \citep{liu2016attention} \\Slot-Gated \citep{goo2018slot} {,} DF-Net \citep{qin-etal-2020-dynamic}
                [BERT: Joint-NLU{,} \citep{chen2019bert} BERT-Joint \citep{castellucci2019multilingual} \\ StackNLU \citep{qin-etal-2019-stack} \\
                RoBERTa: MultiDomain \citep{10.1145/3502198} \\
                Others: MinDS \citep{gerz-etal-2021-multilingual}
                ]
           ]
        ]
        [Dialogue State\\Tracking (DST)
            [RNN: 
            Word-level RNN \citep{henderson-etal-2014-word}{,} NBT \citep{mrksic-etal-2017-neural}{,} MD-DST \\ \citep{mrksic-etal-2015-multi}{,} MemN2N \citep{perez-liu-2017-dialog}{,} GLAD \citep{zhong-etal-2018-global} \\
            GRU: TRADE \citep{wu-etal-2019-transferable}{,} SAS \citep{hu-etal-2020-sas} {,} \\
            LSTM: FJST \citep{eric-etal-2020-multiwoz}{,} 
                [BERT: CSFN \citep{zhu-etal-2020-efficient}{,} SUMBT \citep{lee-etal-2019-sumbt}{,} EMD \citep{wang-etal-2021-fast}{,} \\DS-DST \citep{zhang-etal-2020-find} \\
                Albert: DiCoS \citep{guo-etal-2022-beyond} \\
                GPT-2: NP-DST \citep{ham-etal-2020-end}{,} SimpleToD \citep{SimpleToD}{,} \\ SOLOIST \citep{peng-etal-2021-soloist}
                ]
            ]
        ]
        [Dialogue Policy\\Learning (DPL)
            [RNN: LRU \citep{su2015learning}{,} RS-RNN \citep{su-etal-2015-reward}{,} MADPL \citep{takanobu-etal-2020-multi} \\
            LSTM: OnlineRL \citep{su-etal-2016-line}{,}
            GDPL \citep{gdpl}{,} AdvL \citep{liu-lane-2018-adversarial}
                [BERT: PPO-OFF-Comb \citep{mine_icassp} \\
                GPT2: SimpleToD \citep{SimpleToD}
                ]
            ]
        ]
        [Natural Language\\Generation (NLG)
            [ LSTM: SC-LSTM \citep{wen-etal-2015-semantically}{,} Meta-NLG \citep{10.5555/3367471.3367479}{,} HDSA \citep{chen-etal-2019-semantically}
                [GPT2: SimpleToD \citep{SimpleToD}{,} SC-GPT \citep{peng-etal-2020-shot}]
            ]
        ]
        [End-to-End (E2E)
            [LSTM: Network \citep{wen-etal-2017-network}
                [See Section~\ref{fusion_insides}]
            ]
        ]
    ]
    [ODD
        [ Generative DS
            [ RNN: NRM \citep{weibo}{,} Ext-ED \citep{parthasarathi-pineau-2018-extending} \\
            LSTM: NCM \citep{vinyals2015neural}{,} P-NCM \citep{li2016persona}{,} PersonaChat \citep{personachat} \\
            GRU: CoPerHED \citep{kottur2017exploring}{,} STAR-BTM \citep{zhang2020modeling}{,} \\ CKG-aware \citep{zhong2021keyword}{,} K-NCM \citep{ghazvininejad2018knowledge} \\
                [ Encoder-Decoder: Meena \citep{meena}{,} EVA \citep{zhou2021eva, gu2022eva20}{,} \\ Blenderbot-1{,}2{,}3 \citep{blenderbot,xu-etal-2022-beyond,komeili-etal-2022-internet,shuster2022blenderbot}{,}  \\ PLATO-1{,}2{,}3{,}XL \citep{plato,bao-etal-2021-plato,bao-etal-2022-plato} \\
                GPT2: DialoGPT \citep{zhang-etal-2020-dialogpt}{,} TopicRefine \citep{wang2021topicrefine}{,} \\ CDialoGPT \citep{cdialgpt}{,} Pangu-Bot \citep{mi2022pangubot}
            ]]
        ]    
        [ Retrieval-based DS
            [ RNN: SMN \citep{wu-etal-2017-sequential} \\
              LSTM: DL2P \citep{10.1145/2911451.2911542}{,} Match-LSTM \citep{10.5555/3060832.3061030}{,} \\ Attentive-LSTM \citep{tan2016lstmbased} \\
              GRU: Multi-view \citep{zhou-etal-2016-multi}{,} DUA \citep{zhang-etal-2018-modeling}{,} \\
              CNN: ARC \citep{10.5555/2969033.2969055}{,} RSTP \citep{10.1145/2766462.2767738}{,} LSM \citep{10.1145/2661829.2661935} \\
                [Encoder-Decoder: ConveRT \citep{henderson-etal-2020-convert}{,} WoW \citep{wow} \\
                BERT: Vanilla{,} SA-BERT \citep{gu2020speaker}{,} UMS \citep{whang2021response}{,} \\ SS-DA \citep{lu2020improving}{,} RGD \citep{lin-etal-2020-world}{,} MSN \citep{yuan-etal-2019-multi} \\
                  GPT2: DIALOGRPT \citep{gao-etal-2020-dialogue}]
            ]
        ]
        [ Hybrid
            [ RNN: XiaoIce \citep{zhou2020design}{,} EnsembleGAN \citep{10.1145/3331184.3331193} \\
            GRU: Ensemble \citep{10.5555/3304222.3304379}{,} PTE \citep{10.1609/aaai.v33i01.33017281} \\
            LSTM: EED \citep{pandey-etal-2018-exemplar}{,} RR \citep{weston-etal-2018-retrieve}{,} HybridNCM \citep{10.1145/3357384.3357881} \\
            [
            Encoder-Decoder: DRAME \citep{Technology2019} \\
            GPT2: AdapterBot \citep{madotto2020adapterbot} \\
            ]
            ]
        ]
    ]
    [FuseDS
        [ TOD $\rightarrow$ ODD 
            [ UniDS \citep{zhao2021unids}{,} HyKnow \citep{gao-etal-2021-hyknow}{,} ACCENTOR \citep{sun-etal-2021-adding} FusedChat \citep{DBLP:conf/aaai/YoungXPNC22}, text width=25em 
            ]
        ]
        [ ODD $\rightarrow$ TOD
            [ Q-ToD \citep{q-tod}{,} DuClarifyDial \citep{types_of_dialogs}{,} GODEL \citep{peng2022godel}{,} DialogStudio \citep{zhang2023dialogstudio}{,} MDS \citep{nehring-etal-2021-combining}, text width=25em 
            ]
        ]
    ]
  ]
  \end{forest}
  \caption{Taxonomy of LM-based Dialogue System, with representative LMs/Method of each stage: 1) Independent development of TOD; 2) Independent development of ODD; and 3) Fusion between TOD and ODD. }
  \label{fig:nlm_to_plm}
  \vspace{-6mm}
\end{figure*}

\subsection{The Development of TOD}

In the nascent stages of task-oriented dialogue system development, exemplified by systems like \texttt{GUS} \citep{bobrow1977gus} and \texttt{ALICE} \citep{alice}, the technology relied heavily on rule-based approaches. These early systems operated within single domains and adhered to pre-defined decision trees or rule sets. For example, \texttt{GUS} was constrained to specific domains such as flight information retrieval \citep{bobrow1977gus}, while \texttt{ALICE} was designed to simulate natural language conversation, following a script-like structure based on pattern matching and predefined responses \citep{alice}. These systems showcase impressive capabilities within their limited contexts but struggle to handle nuanced or complex conversations. They often require extensive manual crafting of rules and lack the ability to generalize beyond their narrow domains, making them ill-equipped for real-world applications with varying user intents and open-ended dialogues.

\textit{With the advent of the neural language model (NLM) and subsequent advancements in recurrent neural networks (RNNs)~\citep{lstm, gru}, the landscape of task-oriented dialogue systems underwent a first evolution.} The introduction of RNNs allows for more flexible and adaptive responses, enabling systems to process and generate natural language in a contextually aware manner. Dialogue systems now exhibit a broader spectrum of capabilities, embracing multi-domain functionality and accommodating diverse user intents. Unlike their predecessors, these modern systems can handle complex, intertwined tasks within a single conversation. For instance, a task-oriented dialogue system might seamlessly assist a user in booking a flight, reserving a hotel room, and planning an itinerary, all within the same interaction. This expanded capability has been made possible by the integration of four different components, forming a common pipeline for existing task-oriented dialogue systems: Natural Language Understanding (NLU) \citep{nlu_survey}, Dialogue State Tracking (DST) \citep{dst_survey}, Dialogue Policy Learning (DPL) \cite{mine_mir}, and Natural Language Generation (NLG) \citep{li2021interpretable} \footnote{\url{https://github.com/yizhen20133868/Awesome-TOD-NLG-Survey.git}}. Nevertheless, it is important to note that most studies at this stage focus on individual components and propose task-specific designs, leading to limited scalability and poor generalization. Moreover, there are some works try to combine consecutive components, such as word-level DST (coupling NLU and DST) \citep{henderson-etal-2014-word, mrksic-etal-2017-neural, mrksic-etal-2015-multi, perez-liu-2017-dialog} and word-level policy (coupling Policy and NLG). As shown in Figure~\ref{fig:nlm_to_plm}, the evolutionary trajectory of each sub-module in TOD has been underpinned by the rapid progress of neural language models.

\textit{Notably, a pivotal transformation has been observed in the transition from conventional neural language models (NLMs) to the more versatile pre-trained language models (PLMs).} This shift has significantly contributed to the enhancement of the dialogue system's capabilities and performance. Specifically, there are two obvious trends in this progress. First, TOD initially is designed to perform specific tasks within a predefined domain, however, as the PLMs emerged and gained traction, this paradigm shifted towards \textit{pre-train and then finetune} \citep{pengfeiliu_survey}. By taking advantage of intricate patterns and world knowledge in the parameters of PLMS, the following method based on PLMs can generalize across tasks \citep{peng2017composite} and domains \citep{hakkani2016multi}, even languages \citep{wang2021kddres}. As tasks increase in complexity, they give rise to emerging challenges, including but not limited to insufficient availability of annotated data and the exacerbation of error propagation phenomena. To address these issues, inspired by the notable achievements of PLMs, a considerable body of research endeavors to employ an akin approach to conduct \textit{pre-training or in-domain finetuning on corpora specific to these sub-tasks in TOD simultaneously}, constituting the second discernible trend. For example, SimpleTOD \citep{SimpleToD} uses a single, causal language model trained on all sub-tasks recast as a single sequence prediction problem, leading to better performance at all sub-tasks and robustness to noises. Consequently, these obstacles between distinct components and tasks gradually diminish, and more unified end-to-end TODs are on the way (\S ~\ref{fusion_insides}).

\vspace{-4mm}
\subsection{The Development of ODD}

Distinguishing from TOD, open-domain dialogue systems (ODDs) aim to perform chit-chat with users without task and domain restrictions and are usually fully data-driven in an end2end manner \citep{survey_ds, challenges_in_open_domain}. Generally, there are three types of open-domain dialogue systems: \textit{\textbf{generative dialogue systems}} \citep{li2016persona,zhang2020modeling,wang2021topicrefine}, \textit{\textbf{retrieval-based dialogue systems}} \citep{wu-etal-2017-sequential, tan2016lstmbased, henderson-etal-2020-convert, whang2021response, gao-etal-2020-dialogue}, and \textit{\textbf{hybrid dialogue systems}} \citep{zhou2020design, 10.1145/3331184.3331193, madotto2020adapterbot}. In detail, the retrieval-based systems retrieve predetermined responses from a fixed set in response to user input. In contrast, generative systems construct responses from scratch using language models, resulting in diverse and contextually appropriate outputs. Hybrid systems combine both approaches, using retrieval for efficiency and generative methods for flexibility, aiming to achieve a balance between accuracy and creativity. The choice of system type depends on factors such as the application's requirements, available resources, and the desired user experience. Some representative works are summarized in Figure~\ref{fig:nlm_to_plm}.

As early as 2016, shortly after the introduction of the seq2seq framework, the Alexa Prize challenge was launched to promote the development of open-domain dialogue (ODD) systems capable of engaging in coherent and engaging conversations with users on open-ended topics like sports, politics, and entertainment \citep{chen2018gunrock}. Since then, ODD also went through a shift from NLM to PLM since the appearance of GPT \citep{gpt}, which operates in a decoder-only framework and is trained by predicting the next token auto-regressively. During the era of NLM, most ODD utilizes two typical recurrent neural networks: LSTM \citep{lstm} and GRU \citep{gru} such as response selection tasks \citep{gao-etal-2020-dialogue}, knowledge-grounded dialogue \citep{parthasarathi-pineau-2018-extending, ghazvininejad2018knowledge, wang-etal-2023-retrieval}, persona-consistent dialogue \citep{li2016persona, kottur2017exploring, personachat, persona-consistency}, keywords \citep{zhong2021keyword} and topic \citep{zhang2020modeling} guided dialogue. A similar trend continues except the backbone of the dialogue system becomes PLMs after the appearance of BERT \citep{devlin-etal-2019-bert} and GPT2 \citep{gpt2}. Inspired by the exceptional capability learned during the pre-training stage, lots of works either directly pre-train the model from scratch using the dialogue dataset or further conduct in-domain finetuning using the current PLM as the initial checkpoint. The model after pretraining on tremendous dialogue datasets is formally defined as pre-trained dialogue model \textbf{(DM)}.

There are two common architectures for different PDMs: \textit{encoder-decoder} \citep{meena, gu2022eva20, blenderbot} and \textit{decoder-only} \citep{zhang-etal-2020-dialogpt, cdialgpt}, particularly for generative and hybrid ODD systems. This contrasts with TOD systems, which typically use \textit{encoder-only} PLMs. The distinction here can be attributed to the core nature and objectives of TOD and ODD, as the components of TOD primarily revolve around natural language understanding tasks such as NLU, DST, and DPL, while ODD focuses on natural language generation tasks. Specifically, as shown in the bottom part of Figure~\ref{fig:nlm_to_plm}, Meena \citep{meena} from Google, Blenderbot \citep{blenderbot} from Meta, and EVA \citep{gu2022eva20} from Tsinghua University are all based on encoder-decoder architecture, which demonstrates strong understanding and generation capability simultaneously. In addition, most decoder-only PDMs follow the architecture of GPT2 \citep{gpt2} such as DialoGPT \citep{zhang-etal-2020-dialogpt} and CDialoGPT \citep{cdialgpt} for English and Chinese dialogue respectively. Back to this moment, it is observed that most industrial and academic institutions choose to select decoder-only architecture aiming to improve since the scaling law has been widely observed in causal decoders.


\subsection{Summary}

At this stage, TOD and ODD exhibit distinct boundaries in the data, task and model architectural designs. While TOD focuses on goal-driven interactions (e.g., hotel bookings) with structured pipelines, ODD prioritizes free-form, contextually rich conversations. Advances in pre-trained language models (PLMs) have propelled both paradigms toward more versatile and capable conversational agents. However, their development remains characterized by fragmented methodologies and diverse experimentation, with significant exploration and uncertainty in key areas: i) data: Domain-specific and task-specific corpora dominate TOD, whereas ODD relies on open-domain datasets, results in significant differences in data schema and format between the two; ii) task: TOD modularizes sub-tasks (NLU, DST, NLG), while ODD emphasizes end-to-end generation; and iii) backbone model: most of sub-tasks in TOD adopt encoder-only architectures (e.g., BERT) while ODD favors decoder-only models (e.g., GPT).

Due to the existing complexity and uncertainty, there is no one-size-fits-all solution for TOD and ODD. Fortunately, the advent of PLMs has opened a promising path toward more unified, scalable, and powerful dialogue systems. The development of dialogue systems afterwards is, therefore, an ongoing process aimed at reducing this complexity and uncertainty, ultimately leading to a universal approach that evolves through a bottom-up integration.



\section{Fusion of Dialogue Systems}
\label{fusion_of_ds}

Benefiting from scaling law \citep{kaplan2020scaling}, the performance of pre-trained language models (PLM) can be significantly improved by increasing the model size and training data volume. This scaling enables language models to achieve higher levels of understanding and generation capabilities simultaneously. For example, the BERT \citep{devlin-etal-2019-bert} and GPT series models \citep{gpt} exemplify this capability, leading to three kinds of fusions during the scaling: 1): \textit{fusion within task-oriented dialogue system}; 2) \textit{fusion between two types of dialogue systems}: task-oriented dialogue system and open-domain dialogue system; and 3) \textit{fusion between large language model and dialogue model}.

\vspace{-2mm}
\subsection{Fusion within TOD}
\label{fusion_insides}

Such modular TOD faces several critical limitations, including cascading errors from different components (e.g., NLU misclassifications corrupting the later DST) and high costs for domain-specific adaptation, resulting in low scalability and generalizability. The advert of seq-to-seq framework and especially the appearance of PLM make it possible to unify all sub-tasks into shared architectures, reducing extensive manual feature engineering for each module in traditional TOD. Due to extensive world knowledge pre-trained in PLM, it brings much better scalability, generalizability and also the overall performance.

Specifically, as PLMs demonstrates exceptional performance across various natural language processing downstream tasks, an increasing number of studies seek to leverage them to address multiple sub-tasks in TOD  (e.g., NLU, DST, DPL, and NLG) simultaneously within a unified end-to-end framework \citep{wen-etal-2017-network,moss,peng-etal-2021-soloist,su-etal-2022-multi,he2022galaxy,space-3}, starting from the earliest work NLM (e.g, LSTM, RNN) to the latter PLM (e.g, GPT2, T5), as shown in Table~\ref{tab:e2e_tod}. Several conclusion can be drawn from the results. In general, there is a significant increase in the number of studies aiming to solve multiple sub-tasks in TOD simultaneously \citep{peng-etal-2021-soloist, SimpleToD, 10.1145/3623381}. Some also explore incorporating auxiliary tasks, such as masked language modeling (MLM) \citep{space-3} and response selection (RS) \citep{he2022galaxy}, to enhance overall performance. All of these approaches benefit from the seq-to-seq framework, leveraging various base models based on encoder-decoder or decoder-only architectures.

In detail, the inherent characteristics of each task and the diverse corresponding data formats necessitate distinct strategies for effectively fusing multiple tasks in end-to-end TOD frameworks. On the task level, it is necessary to incorporate the final response generation task (a.k.a, NLG), but the other three sub-tasks are not essential for end-to-end TOD, given that there are multiple methods that do not involve any of the other three tasks. According to the frequency of each task in the different methods, the order relationship is NLG $>$ DST $>$ DPL $>$ NLU. Besides the NLG, DST which tracks the belief state during the conversation becomes more important, since the task-oriented dialogues include many interactions with a specific goal. Furthermore, apart from four key sub-tasks in traditional TOD, we can find more and more additional tasks are optimized with the mentioned key sub-tasks simultaneously. For example, Kulhanek et al. \cite{kulhanek-etal-2021-augpt} use \texttt{BC}, \texttt{SAC}, and \texttt{IC} to further improve the diversity and quality of responses. Similarly, He et al. \citep{space-3} use \texttt{MLM} and \texttt{SRM} to learn from large-scale dialog corpora with limited annotations. On the the data level, SOLOIST \citep{peng-etal-2021-soloist}, as the first work to investigate the effects of pre-training on heterogeneous dialogue corpora, demonstrates great adaptability to accomplish new tasks efficiently. Afterward, lots of studies utilize additional data at the pre-training stage \citep{peng-etal-2021-soloist,kulhanek-etal-2021-augpt,space-3} or fine-tuning stage \citep{su-etal-2022-multi,he2022galaxy}, such as He et al. \citep{he2022galaxy} propose GALAXY to explicitly learn dialog policy from limited labeled dialogs and large-scale unlabeled dialog corpora via semi-supervised learning. The only exception is OPERA \citep{10.1145/3623381}, which proposes a novel dataset OB-MultiWOZ by incorporating question-answering tasks into task-oriented dialogues. 

\begin{table*}[!t]
    \renewcommand{\arraystretch}{1.3}
    \caption{An overview of common E2E Task-oriented Dialogue System, listed in time order.}
    \vspace{-2mm}
    \label{tab:e2e_tod}
    \centering
    \begin{adjustbox}{max width=0.8\textwidth}
    \begin{tabular}{c|c|c|c|c|c|c|c|c|c}
    \hline
    Model & Year & Base Model & NLU & DST & DPL & NLG & Add Tasks* & Add Data & Evaluation Dataset \\
    \hline
    NN \citep{wen-etal-2017-network} & 2017 & LSTM, RNN, CNN & \cmark & \cmark & \cmark & \cmark & \xmark & \xmark &  Wizard-of-Oz Data \\
    MOSS \citep{moss} & 2019 & GRU & \cmark & \cmark & \cmark & \cmark & \xmark  & \xmark & CamRest676, LaptopNetwork \\
    MinTL \citep{mintl} & 2020 & T5, BART & & \cmark & & \cmark & \xmark & \xmark & MultiWOZ 2.0 \\
    SimpleTOD \citep{SimpleToD} & 2020 & GPT2 & 
    & \cmark & \cmark & \cmark & \xmark & \xmark & MultiWOZ 2.0 \\
    SOLOIST \citep{peng-etal-2021-soloist} & 2021 & GPT2 &  & \cmark & & \cmark & \texttt{Cont} & \cmark & CamRest676, MultiWOZ 2.0 \\
    UBAR \citep{yang2021ubar} & 2021 & DistilGPT2 &  & \cmark & \cmark & \cmark & \xmark & \xmark & MultiWOZ 2.0, 2.1 \\
    MTTOD \citep{lee-2021-improving-end} & 2021 & T5 & & \cmark & & \cmark & \texttt{SP} & \xmark &  MultiWOZ 2.0, 2.1 \\ 
    AuGPT \citep{kulhanek-etal-2021-augpt} & 2021 & GPT2 & & \cmark & & \cmark & \texttt{BC}, \texttt{SAC}, \texttt{IC} & \cmark & MultiWOZ 2.0, 2.1 \\
    GALAXY \citep{he2022galaxy} & 2022 & UniLM & & & \cmark & \cmark & \texttt{RS}, \texttt{CR} & \cmark & In-Car, MultiWOZ 2.0, 2.1 \\
    PPTOD \cite{su-etal-2022-multi} & 2022 & T5 & \cmark & \cmark & \cmark & \cmark & \xmark & \cmark & MultiWOZ 2.0 2.1 \\
    OPERA \citep{10.1145/3623381} & 2022 & T5 & & & & \cmark & \texttt{QA} & \xmark & OB-MultiWOZ \\
    SPACE-3 \citep{space-3} & 2022 & UniLM &  \cmark & & \cmark & \cmark & \texttt{MLM},\texttt{SRM} & \cmark & MultiWOZ, CamRest676, In-Car \\
    \hline
    \multicolumn{10}{l}{* Additional task objective abbreviations: \texttt{Cont}-Contrastive Loss, \texttt{SP}--Span Prediction, \texttt{BC}--Belief Consistency, \texttt{SAC}--System
    Action Classification)} \\
    \multicolumn{10}{l}{\texttt{IC}--Intent Classification, \texttt{RS}--Response Selection, \texttt{CR}--Consistency Regularization, \texttt{QA}--Question Answering, \texttt{MLM}-Masked Language Modeling} \\
    \multicolumn{10}{l}{\texttt{SRM}-Semantic Region Modeling} \\
    \end{tabular}
    \end{adjustbox}
\end{table*}

Building on these task- and data-level fusions, the evolution of task-oriented end-to-end dialogue systems also emphasizes more realistic evaluation scenarios. There are some works that recognize the inherent challenge of realistic settings, such as UBAR \citep{yang2021ubar} is evaluated as a fully end-to-end task-oriented dialogue system where its dialog context has access to user utterances and all content it generated such as belief states, system acts, and system responses. Nonetheless, many studies directly assess the performance of different methods either employing various backbones or incorporating additional data \citep{peng-etal-2021-soloist,su-etal-2022-multi,kulhanek-etal-2021-augpt}. Given that the choice of pre-trained language models (PLMs) and the pre-training data significantly impact the final performance, the comparison is not that fair, which underscored the necessity for benchmarking in more realistic and fair settings.

The evolution of task-oriented end-to-end dialogue systems has been marked by significant advancements in data, task and evaluations, which offers notable advantages and also potential limitations. On the positive side, unifying sub-tasks such as NLU, DST, DPL and NLG into a single model reduces cascading errors found in modular pipelines by enabling joint optimization and mitigating the propagation of misalignments between components. Moreover, the unified training format enhances scalability and generalizability, allowing these systems to adapt more effectively to unseen tasks and domains. In contrast, there are still some limitations in terms of data curation and evaluation. To train the model in an end-to-end manner on diverse tasks, it is necessary to maintain a unified data format and pool, leading to increased costs in data curation and annotation. Besides that, evaluating and optimizing such integrated systems becomes more complex since it is difficult to optimize each sub-module independently. Nevertheless, as the understanding and generation capability of language models keep improving with the scaling of data size and model size, the future holds promise for more versatile and effective task-oriented dialogue systems that can understand and respond to user needs in a wide range of domains and languages \citep{mine_mir,systematic_survey}. More details can be found in the corresponding survey paper \citep{qin-etal-2023-end}. 

\subsection{Fusion between TOD and ODD}

The emergence of end-to-end task-oriented dialogue (TOD) systems has catalyzed the development of unified dialogue frameworks capable of integrating both task-oriented and open-domain dialogue (ODD) functionalities. This architectural convergence --- rooted in the adoption of sequence-to-sequence (seq-to-seq) models --- addresses critical limitations of traditional systems while enabling holistic user experiences. Historically, TOD and ODD systems operated in isolation: TOD focused on task completion (e.g., booking flights), while ODD prioritized social interaction (e.g., chit-chat). However, real-world conversations inherently blend these modalities, such as when users discuss travel preferences while completing a booking. Discrete architectures struggled to manage such interleaved interactions, resulting in fragmented user experiences and prohibitive computational costs due to duplicated infrastructure.

As shown in Table~\ref{tab:types_dialogs}, there are different types of dialogue in both TOD and ODD. To seamlessly address them together in an unified manner, lots of work which have embarked on an exploration of integrated task scenarios in order to achieve a more faithful emulation of human-level conversations, where a diverse array of information is seamlessly integrated \citep{types_of_dialogs}. This endeavour extends beyond conventional single dialog types, aiming to integrate ODD and TOD to elevate the authenticity and engagement of dialogue interactions \citep{zhao2021unids,sun-etal-2021-adding,chiu-etal-2022-salesbot,chen-etal-2022-ketod,DBLP:conf/aaai/YoungXPNC22}. This kind of unified dialogue relax the strong assumption that the user must have a clear and specific goal in TOD. Instead, they enable users to refine or discover objectives through multi-turn natural language exchanges, akin to human conversations \citep{types_of_dialogs}. Table~\ref{tab:tod_odd} shows the high-level fusion of TOD and ODD. In general, there are two paths to build a unified dialogue system: expanding TOD to incorporate open-ended dialogue or enhancing ODD with task-oriented capabilities.

\begin{table*}[!t]
    \renewcommand{\arraystretch}{1.0}
    \caption{High-level of fusion between TOD and ODD, wherein there are different types of ODD.}
    \label{tab:tod_odd}
    \centering
    \begin{adjustbox}{max width=0.8\textwidth}
    \begin{tabular}{c|c|c|c|c|c|c}
    \hline
    \multirow{2}{*}{Model} & \multirow{2}{*}{Task}  & \multirow{2}{*}{Type of ODD} & \multicolumn{3}{c|}{Knowledge sources} & \multirow{2}{*}{Training method} \\
    \cline{4-6} & &  & Database & Web & PLMs & \\
    \hline
    Q-TOD \citep{q-tod} & TOD & \xmark & \cmark & \xmark & \xmark & End-to-end \\
    \hline
    Internet-Augmented \cite{komeili-etal-2022-internet} & ODD & KG Chit-Chat & \xmark & \cmark & \xmark & End-to-end  \\
    \hline
    \multirow{2}{*}{UniDS \citep{zhao2021unids}, FusedChat \citep{DBLP:conf/aaai/YoungXPNC22}, SalesBot \citep{chiu-etal-2022-salesbot}} & \multirow{2}{*}{TOD + ODD} & \multirow{2}{*}{Chit-Chat} & \multirow{2}{*}{\cmark} & \multirow{2}{*}{\xmark} & \multirow{2}{*}{\xmark} & \multirow{2}{*}{End-to-end} \\
    
     &  & & & & \\
     
    \hline
    HyKnow \citep{gao-etal-2021-hyknow}, KETOD \citep{chen-etal-2022-ketod} & TOD + ODD & KG Chit-Chat & \cmark & \cmark & \xmark & End-to-end \\
    \hline
    GODEL \citep{peng2022godel} & TOD + ODD & Question Answering, KG Chit-Chat & \cmark & \cmark & \xmark & End-to-end  \\
    DuClarifyDial \citep{types_of_dialogs} &  TOD + ODD & Chit-Chat, Question Answering, KG Chit-Chat & \cmark & \cmark & \xmark & End-to-end \\
    MDS \citep{nehring-etal-2021-combining} & TOD + ODD & Question Answering & \cmark & \cmark & \xmark & End-to-end \\
    OPERA \citep{10.1145/3623381} & TOD + ODD & Question Answering & \cmark & \cmark & \cmark & End-to-end \\
    DialogStudio \citep{zhang2023dialogstudio} & TOD + ODD & Dial-Sum, NLU, Conv-Rec, KG Chit-Chat & \cmark & \cmark & \xmark & End-to-end \\
    \hline
    \end{tabular}
    \end{adjustbox}
    \vspace{-4mm}
\end{table*}

The first path, extending TOD to ODD, involves taking existing task-oriented dialogue systems and expanding their capabilities to handle more open-ended, free-form conversation. Specifically, since TOD consists of unique data schema such as belief state and database results, some of previous methods attempt to build a unified data schema to making it can extend to open-domain settings, falling into the first category.  Lots of other works conceptualize open-domain dialogue as an end-to-end response generation task, where external knowledge is treated as a special type of database result \citep{zhao2021unids, sun-etal-2021-adding,DBLP:conf/aaai/YoungXPNC22,types_of_dialogs}. For examples, both UniDS \citep{zhao2021unids} and PLATO-MT \citep{types_of_dialogs} choose to optimize three sub-tasks: belief state generation, system act generation, and response generation, in an end-to-end manner. Moreover, \citet{DBLP:conf/aaai/YoungXPNC22} further feature inter-mode contextual dependency between TOD and ODD, i.e., the dialogue turns from the two modes depending on each other. To conclude, the unified data schema usually contains \textit{(user input, belief state, DB result, system act, system response)} following the traditional TOD setting. Instead of proposing an unified schema, a simpler approach is to directly insert additional open-domain dialogues into task-oriented dialogues. In this way, the system responds based on knowledge from domain FAQs \citep{kim-etal-2020-beyond} or unstructured documents \citep{gao-etal-2021-hyknow} whenever it detects that the current query requires open-domain knowledge. These datasets mentioned primarily address the task of identifying conversational turns that necessitate external knowledge for generating responses and selecting the pertinent knowledge for response construction.

Conversely, the second path, enhancing ODD with TOD capabilities, focuses on equipping open-domain systems with structured and task-oriented interactions. This involves adding the ability to understand user intents, track dialogue state, and complete well-defined tasks, on top of the open-ended conversational capabilities. Previous researchers either use a retrieval-augmented framework \citep{q-tod, peng2022godel, zhang2023dialogstudio}, treating external databases (in TOD) or documents (in ODD) as special sources of knowledge, or they modularize each component within the dialogue systems \citep{10.1145/3623381, zhang2023dialogstudio}. Taking Q-ToD as an example, \citet{q-tod} introduce a novel query-driven task-oriented dialogue system, namely Q-TOD. The essential information from the dialogue context is extracted into a query, which is further employed to retrieve relevant knowledge records for response generation, following the conventional architecture of open-domain dialogue systems. Similarly, \citet{peng2022godel} utilize a similar architecture but focus on the effects of pretraining with a comprehensive dialogue corpus. Furthermore, several attempts apply the modular dialog system framework to combine TOD and ODD. For example, \citet{nehring-etal-2021-combining} propose a hierarchical architecture composed of several dialogue systems, and then use a classifier to select the right routine to answer current user utterance. We emphasise that both of these two paths share a common objective which is to intertwine these two dialogue modes together seamlessly in the same conversation.

Beyond fusing TOD and ODD, researchers try to  incorporate additional skills, tasks, and knowledge sources, enriching the scope and boundaries of dialogue system, making them more adaptable and human-like. i) \textit{Skills}. \citet{madotto2020adapterbot} introduce the Adapter-Bot, a dialogue model that uses a fixed backbone conversational model such as DialGPT \citep{zhang-etal-2020-dialogpt} and triggers on-demand dialogue skills (e.g., emphatic response, weather information, movie recommendation) via different adapters, while \citet{smith-etal-2020-put} also try to blend multiple skills into one conversation session, in which knowledge, emotional and personalizing skills are shown together in the same multi-turn conversation. Some recent studies focus on other skills of dialogue system, such as proactivity \citep{proactive_ds_survey} and different conversational strategies \citep{wang2023tpe}; ii) \textit{Tasks.} Besides the combination of question-answering and TOD tasks (as mentioned in \S~\ref{fusion_insides}), DuClarifyDial \citep{types_of_dialogs} additionally incorporate chit-chat, knowledge-grounded chit-chat while GODEL \citep{peng2022godel} specifically target knowledge-grounded goal-directed dialogs by leveraging a new phase of grounded pre-training on open-domain goal-directed corpus. There is a latest work, DialogStudio, which further encompasses data from natural language understanding, conversation recommendation, and dialogue summarization \citep{zhang2023dialogstudio}. iii) \textit{Knowledge Sources.} Most of the previous works either rely on the dataset knowledge for TOD or the Web document for ODD \citep{q-tod,komeili-etal-2022-internet}, and recent works generalize the possible knowledge sources to include all databases (i.e., user memory, domain-specific database), the Web, and even pre-trained language model itself at the same time \citep{10.1145/3623381}. More analysis can be found in \S~\ref{external_llm_ds}.

The integration of TOD and ODD into unified frameworks offers significant benefits but also introduces new complexities. On the positive side, these systems enhance the user experience by seamlessly interleaving task completion (e.g., booking a hotel) with social interaction (e.g., discussing travel preferences), thereby eliminating the jarring transitions seen in modular systems. This unified approach not only paves the way toward more human-like conversational agents but also improves scalability through joint optimization of diverse sub-tasks. However, the fusion of these dialogue modes presents notable challenges. Unified models require large-scale, complex annotated datasets to accommodate varied data formats and tasks, which significantly increases the cost and difficulty of data curation. Moreover, blending distinct dialogue modalities complicates error analysis, making it more challenging to identify and correct issues. Finally, the inherent trade-off between generality and specificity may lead to performance gaps, where enhancing one modality could inadvertently compromise another, ultimately affecting the system's overall reliability and interpretability.

\subsection{Fusion between LLM and DM}

With the continual influx of data, model sizes have grown significantly at the same time (i.e., the scaling law~\citep{kaplan2020scaling}), leading to the appearance of large language models (LLMs)~\citep{instructgpt,touvron2023llama2}. These LLMs exhibit exceptional generalization across diverse tasks while minimizing the need for extensive task-specific fine-tuning that traditional dialogue models require. Moreover, the universal training recipe, including instruction-tuning and reinforcement learning from human feedback (RLHF), empower LLMs to adapt flexibly to conversational contexts, bridging the gap between open-domain chit-chat and goal-oriented dialogue.

This shift was catalyzed in 2022 by OpenAI’s groundbreaking product --- ChatGPT \footnote{https://chatgpt.com/} -- a single model that seamlessly integrates contextual understanding, instruction following, and task completion to deliver responses that are both helpful and safe. Consequently, the boundaries between task-oriented and open-domain dialogues have become increasingly indistinct, since LLMs are beginning to demonstrate a preliminary capability for both task-oriented \citep{zhang2023sgptod} and all other types of dialogues \citep{wang2023emotional} even under the zero-shot or few-shot settings. Following the release of ChatGPT, numerous LLMs are proposed and corresponding dialogue models (DM) are finetuned on different-sized dialogue corpus, aiming to achieve better performance on different dialogue tasks, domains or even languages \citep{du2022glm, touvron2023llama2}. Hence, there has been an exponential growth in both LLMs and DMs, encompassing both open-source and proprietary (closed-source) models. To provide a more comprehensive analysis and comparison of advancements, we focus on the two most widely used languages --- Chinese and English --- before discussing multilingual LLMs in the final part. Table~\ref{tab:dm_llm} shows the overview of dialogue model based on different base models.

\textbf{LLMs and DMs in English.} Out of lots of English models, Meta’s LLaMA series \citep{touvron2023llama,touvron2023llama2} is a leading example of open-sourced LLMs, while OpenAI’s GPT series \citep{instructgpt,gpt4} represents one of the most well-known proprietary models. On the one hand, the evolution of the GPT-series models initiated with GPT \citep{gpt} and has been consistently refined by OpenAI. This improvement involves scaling both the model size and the pre-training data size, including the incorporation of code data \citep{chen2021evaluating} and increasing the model parameters to a larger size, notably reaching 175 billion in GPT-3 \cite{brown2020language}. The culmination of these enhancements is evident in the success of ChatGPT, achieved through additional fine-tuning and alignment techniques from human feedback \citep{ouyang2022traininglanguagemodelsfollow} based on the GPT-3.5 model. Initially trained on a dataset of 570GB, ChatGPT supports input lengths of 4k and 16k, and demonstrates remarkable capabilities in human communication, boasting a substantial knowledge base, proficiency in reasoning through mathematical problems, accurate contextual understanding in multi-turn dialogues, alignment with human values to ensure safe use and following diverse user instructions. Subsequently, the introduction of GPT4 \citep{gpt4} further extends the model's capabilities, enabling the resolution of multi-modal tasks involving both image and text inputs and outputs. It is worthy noting that the maximum input window size is also doubled, which empowers the model to handle long context input. Benefiting from the cooperation with OpenAI, Microsoft release New Bing\footnote{\url{https://blogs.microsoft.com/wp-content/uploads/prod/sites/5/2023/02/The-new-Bing-Our-approach-to-Responsible-AI.pdf}}, which is a conversational agent based on GPT4, capabling to provide up-to-date information with the assistance of search engine. 

\begin{table*}[]
    \renewcommand{\arraystretch}{1.3}
    \caption{An overview of Dialogue Models based on Large Language Models listed in time order.}
    \label{tab:dm_llm}
    \begin{adjustbox}{max width=0.8\textwidth}
    \begin{tabular}{c|c|c|c|c|c|c|c}
    \hline
    Dialogue Models & Release Date & Base Model              & \# of Params   & Window size & Trainig Tokens & Language Preference & Source Type \\ \hline
    BlenderBot3\citep{shuster2022blenderbot}   & Aug, 2022     & OPT                     & 3B/30B/175B    &  \xmark           & 180B           & EN             & Open        \\
    ChatGPT       & Dec, 2022     & GPT3.5 \citep{brown2020language}                 & 175B           & 4k/16k        &  \xmark              & EN             & Closed      \\
    Bing Chat      & Feb, 2023     & GPT4 \citep{gpt4}                   &  \xmark              & 8k/32k        &  \xmark             & EN             & Closed      \\
    LLaMA-Chat \citep{touvron2023llama}    & Feb, 2023     & LLaMA    & 7B/13B/33B/65B & 2k        & 1.4T           & EN             & Open        \\
    Alpaca \citep{alpaca}        & Mar, 2023     &   LLaMA                      & 7B             & 2k        & 1.4T           & EN             & Open        \\
    Vicuna \citep{chiang2023vicuna}        & Mar, 2023     &  LLaMA                          & 7B/13B        & 2k        & 1.4T           & EN             & Open        \\
    ChatGLM \citep{zeng2022glm}      & Mar, 2023     & GLM                     & 6B/12B         & 2k         & 1T             & CN             & Open        \\
    Slack      & Mar, 2023     & Claude                  &  \xmark              & 200k        & 2.0T           & EN             & Closed      \\
    Bard         & Mar, 2023     & PALM                    & 540B           &  \xmark           &  \xmark             & EN             & Closed      \\
    
    Falcon-Chat \citep{penedo2023refinedweb}   & Jun, 2023     & Falcon                  & 7.5B/40B/180B  & 2k        & 1.5T/1T/3.5T & EN             & Open        \\
    
    MPT-Chat \citep{lin2023mpt}      & Jun, 2023     & MPT                     & 7B/30B         & 8k          & 1T             & EN             & Closed      \\
    
    InternLM-Chat \citep{2023internlm} & Jun, 2023     & InternLM                & 7B/20B         & 8k          & 2.3T           & CN             & Open        \\
    
    Baichuan-Chat \citep{yang2023baichuan}, & Jun, 2023     & Baichuan                & 7B/13B/53B         & 4k          & 1.2T           & CN             &  Open (7B/13B)     \\
    
    ChatGLM2 \citep{zeng2022glm}      & Jun, 2023     & GLM                     & 6B/12B         & 32k         & 1T             & CN             & Open        \\
    
    LLaMA2-Chat \citep{touvron2023llama2}   & Jul, 2023     & LLaMA2 & 7B/13B/34B/70B & 4k        & 2.0T           & EN             & Open        \\
    
    Atom-Chat    & Aug, 2023     &   LLaMA2                      & 7B/13B         & 4k        & 1.0T           & CN             & Open        \\
    
    Qwen-Chat \citep{bai2023qwen}     & Aug, 2023     & QWen                    & 7B/14B         & 8k          & 2.4T           & CN             & Open        \\ 
    
    Baichuan2-Chat \citep{yang2023baichuan} & Sep, 2023  & Baichuan2    & 7B/13B & 4k & 2.6T   & CN   & Open \\
    
    ChatGLM3 & Oct, 2023  &  GLM  & 6B  & 32k  & /  & CN & Open  \\    
    \hline
    \end{tabular}
    \end{adjustbox}
\vspace{-4mm}
\end{table*}

On the other hand, unlike the closed-source nature of the GPT series models from OpenAI, Meta has embraced openness by releasing the LLaMA \citep{touvron2023llama} and LLaMA2 \citep{touvron2023llama2}, which are pre-trained on 1.4T data, offers four parameter configurations: 7B, 13B, 33B, and 65B. This provides a fresh opportunity for an open-source community, with corresponding dialogue models being consistently finetuned based on LLaMA. LLaMA-Chat, for example, finetunes the LLaMA model using instructional methods, aligning with the GPT-3 autoregressive paradigm. In addition, Alpaca \citep{alpaca} is finetuned by Stanford based on LLaMA-7B, using \texttt{text-davinci-003} API and self-instruct techniques to create a prompt-response instruction dataset of 52K. Vicuna \citep{chiang2023vicuna}, based on LLaMA-13B, used supervised data from ShareGPT \footnote{\url{https://sharegpt.com/}}, extending sequence length from 512 to 2048 during training. To further improve the window size and human values alignment with the public, LLaMA2 \citep{touvron2023llama2} employs grouped-query attention, extending input length from 2k to 4k and training data from 1.4T tokens to 2.0T tokens. Besides these two series, there are other series English LLMs such as Claude\footnote{\url{https://claude.ai/}}, Bard, Falcon, etc. These models typically adhere to an architecture resembling that of GPT-3, showcasing proficiency in handling multi-turn and complex dialogues, with different sizes of input window. Other commercial companies have also adopted similar training methods and subsequently introduced their own conversational large language models, such as MPT-Chat \citep{lin2023mpt} and Falcon-Chat \citep{penedo2023refinedweb}.

\textbf{LLMs and DMs in Chinese.} Besides English, there are also lots of impressive Chinese models. For example, ChatGLM \cite{du2022glm, zhang2023glm}, as the first open-source Chinese LLM, which is a bilingual model (Chinese and English) built on the general language model architecture -- GLM 130B \cite{zhang2023glm}. The open-sourced 6B version, ChatGLM2-6B, extends the context length from 2k to 32k tokens and employs multi-query attention for more efficient inference. Furthermore, they progress from ChatGLM and ChatGLM2 to ChatGLM3 \footnote{\url{https://github.com/THUDM/ChatGLM3/}}, which incorporates a broader and more diverse corpus of data during its fine-tuning process. Additionally, ChatGLM3 has demonstrated significant improvements in planning and reasoning, akin to language agents \citep{zeng2023agenttuning, sumers2023cognitive}, thus reflecting substantial improvements in tool learning, code execution, database operation, knowledge graph search and reasoning. There are publicly available LLMs from other companies, such as Baichuan-Chat \citep{yang2023baichuan} and Baichuan2-Chat. The latter, Baichuan2-Chat, employs a method similar to Claude and is trained on over 2.6T tokens. These models cater to a wide spectrum of applications, each offering unique advantages stemming from their fine-tuning strategies and underlying architecture. Alibaba releases Qwen-7B-Chat aligns with human intent using a curated dataset, pre-trained on over 2.2T tokens with a context length of 2048. Sensetime publish InternLM-Chat, which is derived from supervised fine-tuning and reinforcement learning with human feedback on InternLM \citep{2023internlm}, a multilingual model pre-trained on 2.3T tokens.

\textbf{LLMs and DMs in Other Languages.} Since most of existing LLMs naturally support multiple languages due to comprehensive corpus in different languages involved during pre-training, there is only few attempts to further finetune dialogue model for specific languages, especially for underrepresented languages. We summarize several such attempts for Korean, Sherkala, and more boarder languages at the bottom part of Table~\ref{tab:multilingual_llm}. Specifically, most studies directly finetune the large language model (i.e., QWen and LLaMA) on lots of multilingual corpus to get the multilingual versions. For examples, Suzume 8B~\citep{devine2024tagengo} is a multilingual model fine-tuned on nearly 90,000 conversations using Llama 3, endowing it with Llama 3’s capabilities alongside enhanced multilingual support. In contrast, Panges-7B~\citep{yue2024pangeafullyopenmultilingual} extends this functionality by incorporating multimodal capabilities in addition to multilingual features.

\begin{table*}[]
    \renewcommand{\arraystretch}{1.3}
    \caption{Some of representative multilingual large language models.}
    \label{tab:multilingual_llm}
    \begin{adjustbox}{max width=0.8\textwidth}
    \begin{tabular}{c|c|c|c|c|c|c|c}
    \hline
    Models & Release Date & Base Model              & \# of Params   & Window size & Trainig Tokens & Language Preference & Source Type \\ \hline

    Llama-2-ko-7B-Chat & Dec 2023 & LLaMA2 & 7B & 2k & / & Korean & Open \\

    Suzume 8B~\citep{devine2024tagengo} & May 2024 & LLaMA3 & 8B & 8k & / & Multilingual & Open \\
    
    LLaMAX3-8B~\citep{lu-etal-2024-llamax} & July 2024 & LLaMA3 & 8B & 8k & / & Multilingual & Open \\

    Pangea-7B~\citep{yue2024pangeafullyopenmultilingual} & Oct 2024 & QWen2 & 7B & 32k & / &  Multilingual & Open \\

    Aya-Expanse~\citep{dang2024ayaexpansecombiningresearch} & Dec 2024 & Command & 8B & 8k & / &  Multilingual & Open \\

    Babel-Chat~\citep{zhao2025babelopenmultilinguallarge} & Feb 2025 & QWen & 9B/83B & 131K & / & Multilingual & Open \\
    
    Llama-3.1-Sherkala-8B-Chat~\citep{koto2025sherkala} & Mar 2025 & LLaMA3.1 & 8B & 8k & / & Sherkala & Open \\

    \hline
    \end{tabular}
    \end{adjustbox}
\end{table*}

To conclude, several trends can be drawn from the fusion results at Table~\ref{tab:dm_llm}. First of all, there is a discernible temporal trend towards an increased prevalence of open-sourced LLMs, particularly within the domain of dialogue models. Secondly, this trend is accompanied by continuous growth in both training data and parameter configurations, although most released models tend to be around 7B or 13B parameters due to the high computational costs associated with inference. In addition, there is a noteworthy extension in the window length, denoting an enhanced ability to process and generate longer sequences of text. To further improve the utility and safety of dialogue models, extensive instruction fine-tuning using conversational data is increasingly integrated, and their multilingual capabilities continue to evolve.

\begin{table*}[t]
\centering
\scriptsize
\caption{Performance Comparison of Different Dialogue Methods across TOD and OOD Tasks. \textit{Note:} Blue-highlighted rows indicate \textbf{Unified Dialogue Systems}; yellow-highlighted rows indicate \textbf{End-to-End Dialogue Models}.}
\label{evaluation_results_ds}
\begin{tabular}{ll l l >{\centering\arraybackslash}m{3cm} l}
\toprule
\textbf{Task Type} & \textbf{Dataset} & \textbf{Method Type} 
    & \textbf{Method} & \textbf{Metric} & \textbf{Performance} \\
\midrule

\multirow{4}{*}{ToD-NLU}
 & \multirow{4}{*}{MSDialog \citep{qu2018analyzing}}
   & \multirow{2}{*}{Tra. DS}
     & BERT-base \citep{devlin-etal-2019-bert} & \multirow{4}{*}{P, F1-Micro, F1-Macro} & 81.0, 83.1, 74.3 \\
 &  &  & T5-base &                      & 83.9, 86.8, 77.1 \\
 \cmidrule(lr){3-4}\cmidrule(lr){6-6}
 &  & \multirow{1}{*}{LLM-Prompting}
     & GPT-4o &                        & 54.6, 95.8, 42.4 \\
 \cmidrule(lr){3-4}\cmidrule(lr){6-6}
 &  & \multirow{1}{*}{LLM-Finetuning}
     & SOLID \citep{askari2024self} &                        & 84.2, 87.9, 78.6 \\
\midrule

\multirow{9}{*}{ToD-DST}
 & \multirow{9}{*}{Multi-Woz 2.1 \citep{eric-etal-2020-multiwoz}}
   & \multirow{3}{*}{Tra. DS}
     & \cellcolor{yellow!20}SimpleTOD \citep{SimpleToD} & \multirow{9}{*}{JGA} & 54.15 \\
 &  &  & \cellcolor{yellow!20}TOATOD \citep{bang2023task}  &    & 54.79 \\
 &  &  & DSGFNet \citep{feng2022dynamic}                    &    & 56.79 \\
 \cmidrule(lr){3-4}\cmidrule(lr){6-6}
 &  & \multirow{3}{*}{LLM-Prompting}
     & GPT-4o         &   & 62.59 \\
 &  &  & Llama2-13B   &   & 49.28 \\
 &  &  & VICUNA-13B-V1.5 & & 55.41 \\
 \cmidrule(lr){3-4}\cmidrule(lr){6-6}
 &  & \multirow{3}{*}{LLM-Finetuning}
     & FNCTOD \citep{li2024large}   &    & 59.54 \\
 &  &  & SERIDST \citep{lee2024inference} &    & 56.96 \\
 &  &  & GPT-3.5 &    & 61.31 \\
\midrule

\multirow{8}{*}{ToD-DPL}
 & \multirow{8}{*}{Multi-Woz 2.0 \citep{budzianowski2018multiwoz}}
   & \multirow{2}{*}{Tra. DS}
     & \cellcolor{yellow!20}GALAXY \citep{he2022galaxy}   & \multirow{8}{*}{Inform, Comb} & 93.5, 110.4 \\
 &  &  & DiactTOD \citep{wu2023diacttod} &                   & 94.6, 97.1 \\
 \cmidrule(lr){3-4}\cmidrule(lr){6-6}
 &  & \multirow{1}{*}{Unified DS}
     & \cellcolor{cyan!20}UniDS \citep{zhao2021unids} & & 90.3, 104.12 \\
 \cmidrule(lr){3-4}\cmidrule(lr){6-6}
 &  & \multirow{3}{*}{LLM-Prompting}
     & GPT-4       & & 85.2, 79.1 \\
 &  &  & Llama2 70B & & 54.3, 38.9 \\
 &  &  & Llama2 13B & & 37.1, 33.6 \\
 \cmidrule(lr){3-4}\cmidrule(lr){6-6}
 &  & \multirow{2}{*}{LLM-Finetuning}
     & AutoTOD \citep{xu2024rethinking} & & 85.2, 79.1 \\
 &  &  & GPT-3.5 & & 62.5, 52.7 \\
\midrule

\multirow{7}{*}{ToD-NLG}
 & \multirow{7}{*}{Multi-Woz 2.1}
   & \multirow{2}{*}{Tra. DS}
     & \cellcolor{yellow!20}UBAR \citep{yang2021ubar}    & \multirow{7}{*}{SUCCESS, BLEU} & 89.1, 14.8 \\
 &  &  & \cellcolor{yellow!20}GALAXY \citep{he2022galaxy} &               & 84.9, 20.8 \\
 \cmidrule(lr){3-4}\cmidrule(lr){6-6}
 &  & \multirow{1}{*}{Unified DS}
     & \cellcolor{cyan!20}HyKnow \citep{gao-etal-2021-hyknow} &  & 69.4, 18.0 \\
 \cmidrule(lr){3-4}\cmidrule(lr){6-6}
 &  & \multirow{3}{*}{LLM-Prompting}
     & GPT-3.5 &  & 82.8, 9.3 \\
 &  &  & GPT-4  &  & 84.4, 10.4 \\
 &  &  & Llama2-70B & & 69.8, 7.8 \\
 \cmidrule(lr){3-4}\cmidrule(lr){6-6}
 &  & \multirow{1}{*}{LLM-Finetuning}
     & AutoTOD \citep{xu2024rethinking} &  & 84.9, 7.8 \\
\midrule

\multirow{7}{*}{ToD-E2E}
 & \multirow{7}{*}{Multi-Woz 2.0}
   & \multirow{2}{*}{Tra. DS}
     & \cellcolor{yellow!20}PPTOD \citep{su2021multi} & \multirow{7}{*}{BLEU, Inform, Success} & 18.1, 62.9, 79.4 \\
 &  &  & \cellcolor{yellow!20}SPACE-3 \citep{space-3} & & 19.3, 95.3, 88.0 \\
 \cmidrule(lr){3-4}\cmidrule(lr){6-6}
 &  & \multirow{1}{*}{Unified DS}
     & \cellcolor{cyan!20}UniDS \citep{zhao2021unids} &  & 18.72, 80.3, 80.05 \\
 \cmidrule(lr){3-4}\cmidrule(lr){6-6}
 &  & \multirow{3}{*}{LLM-Prompting}
     & GPT-3.5 &  & 19.4, 87.0, 79.1 \\
 &  &  & GPT-4  &  & 10.4, 91.7, 84.4 \\
 &  &  & Llama2-70B & & 7.8, 73.3, 69.9 \\
 \cmidrule(lr){3-4}\cmidrule(lr){6-6}
 &  & \multirow{1}{*}{LLM-Finetuning}
     & InstructTODS \citep{chung2023instructtods} &  & 9.34, 90.7, 76.2 \\
\midrule

\multirow{5}{*}{OOD}
 & \multirow{5}{*}{ESConv \citep{liu2021towards}}
   & \multirow{2}{*}{Tra. DS}
     & KEMI \citep{deng2023knowledge}     & \multirow{5}{*}{BL-2, RG-L, Dist-3} & 2.7, 2.9, 2.3 \\
 &  &  & TransESC \citep{zhao2023transesc} &                  & 2.6, 2.8, 2.5 \\
 \cmidrule(lr){3-4}\cmidrule(lr){6-6}
 &  & \multirow{2}{*}{LLM-Prompting}
     & Llama-3-70B    & & 2.1, 3.0, 2.6 \\
 &  &  & GPT-3.5-turbo & & 3.4, 2.5, 3.5 \\
 \cmidrule(lr){3-4}\cmidrule(lr){6-6}
 &  & \multirow{1}{*}{LLM-Finetuning}
     & COOPER \citep{cheng2024cooper} & & 2.9, 5.4, 3.4 \\
\bottomrule
\end{tabular}
\end{table*}

The fusion of LLMs and DMs offers transformative benefits, particularly in enhancing performance and improving generalization. Table~\ref{evaluation_results_ds} presents the performance of different types of dialogue system on some representative benchmarks. There are several trends can be drawn from the results. First of all, Traditional task-specific methods (Tra. DS) generally perform well in specific tasks, but they often lack flexibility and adaptability across different dialogue challenges. Secondly, recent LLMs, such as GPT-3.5 and GPT-4, achieve exceptional performance and, in some cases, even outperform traditional task-specific methods (Tra. DS) in a zero-shot setting, in both TOD or ODD tasks like ToD-DST, ToD-E2E and ESConv. Thirdly, some finetuned small sized LLMs can outperform relatively larger sized LLMs on certain tasks, such as AutoToD outperforms GPT-3.5 in ToD-DPL. Overall, by leveraging extensive pretraining, LLMs develop vast world knowledge, nuanced reasoning abilities, and robust generalization across diverse natural language tasks. This comprehensive pretraining significantly reduces the need for task-specific fine-tuning --- not only in dialogue systems but across nearly all language-based applications --- enabling these models to function more like human assistants rather than solely dialogue systems. However, this kind of fusion is not without limitations. The overall performance of an integrated dialogue model is heavily contingent upon the underlying LLM; thus, any biases or limitations inherent in the base model can be transferred to the dialogue system. Moreover, language dominance exacerbates existing inequities: models for widely represented languages like English and Chinese benefit from abundant training data, whereas underrepresented languages (e.g., Sherkala) struggle with sparse data and limited fine-tuning efforts. This imbalance risks cultural homogenization, as dominant languages increasingly shape dialogue norms and priorities. Ultimately, while the fusion of LLMs and DMs marks a significant leap forward, addressing these challenges is essential to ensure that dialogue systems are both equitable and culturally diverse.

\subsection{Summary}

Over the decades, dialogue systems have evolved from rule-based and modular architectures --- where separate components handled tasks like natural language understanding, dialogue management, and response generation --- to unified end-to-end systems that leverage the power of deep learning and large-scale pretraining. The three key waves of fusions, starting from fusions within TOD, then fusion between TOD and ODD, and finally LLM with the DM, exactly reflect the fusion from bottom to top, being more and more scalable and generaliable across data, tasks and models. One of key lessons from the history of dialogue system is exactly \textit{the Bitter Lesson}\footnote{http://www.incompleteideas.net/IncIdeas/BitterLesson.html}, which progress has been driven by scaling computation and data, not human-centric engineering, less structure, more intelligence. These difference levels of fusions can be regarded as a progress keeping reduce human-centric engineering but leverage scaling computation and data.

Looking forward, the future of dialogue systems will likely continue to be driven by the bitter lesson’s core tenet: computation and learning, not human ingenuity, are the engines of intelligence. As models become increasingly unified --- blurring the lines between different tasks, languages and modalities -- they promise more natural, context-aware, and adaptable interactions. However, these new fusions also bring challenges, including the new learning paradigm, unified data format and even new model architecture as we observed in the development of dialogue system. Ultimately, by embracing the lessons of the past and leveraging the immense power of computation, the next generation of dialogue systems can become more robust, versatile, and human-like in their understanding and interactions.

\section{LLM-based Dialogue System}
\label{llm-based_ds}

Nowadays, it is not a formidable challenge for LLM-based dialogue systems to pass the esteemed \textit{Turing Test}~\citep{jones2025largelanguagemodelspass}. More importantly, as dialogue systems and backbone language models evolve, the boundaries between tasks, models, and even languages are likely to blur. A fundamental capability of such models will be their ability to mimic human cognitive processes and interact effectively with external environments, in order to get more helpful, harmless and natural human-like response. This raises the critical question of how models can employ cognitive reasoning internally and simultaneously interact with environment externally. The unification of these two abilities will be very foundation to creating models that not only reason current state but also act as adaptive problem-solvers in complex real-world contexts.

\begin{figure}[!t]
    \centering
    \includegraphics[trim={5.5cm 4.5cm 4cm 3cm}, clip, scale=1.0, width=0.8\textwidth]{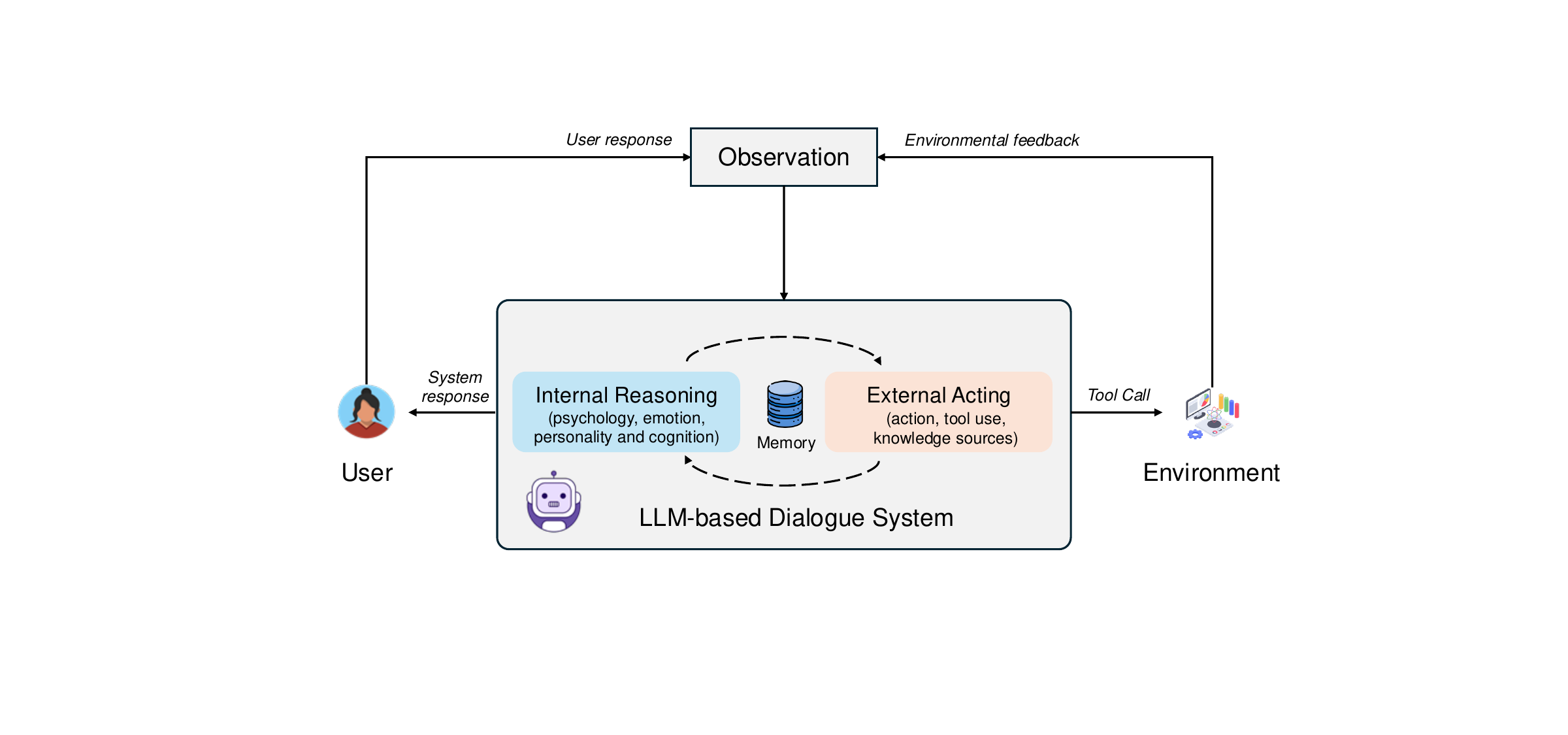}
    \caption{A novel unified perspective of LLM-based Dialogue System. There are three conversational flows: 1) response based on internal reasoning and dialogue context; 2) response based on external acting and dialogue context; and 3) response based on internal reasoning, external acting, and dialogue context.}
    \label{fig:new_problems_llm_ds}
    \vspace{-4mm}
\end{figure}

Figure~\ref{fig:new_problems_llm_ds} shows the three major conversational flows for the LLM-based dialogue system. First of all, as shown in left side, the LLM-based dialogue system need to understand the user intention and interprets subtle cues or implicit expectations with greater nuance by leveraging exceptional reasoning capabilities of LLMs (\S~\ref{internal_llm_ds}), similar with the recently released GPT 4.5\footnote{https://openai.com/index/introducing-gpt-4-5/}. Secondly, it should interact with various external knowledge sources --- such as long-term memory systems, private databases, and domain-specific services --- to generate more informed and accurate responses with reduced hallucinations and up-to-date information (\S~\ref{external_llm_ds}). Lastly, the system carries out these operations through a multi-step process, enabling more sophisticated decision-making and problem-solving (\S~\ref{internal_exteal_llm_ds}). Together, these three critical capabilities have become increasingly attractive and crucial for empowering dialogue systems, which can be utilized in complementary to suit the diverse and complicated needs in practice.

\subsection{Internal Reasoning}
\label{internal_llm_ds}

\begin{table*}[!t]
    \renewcommand{\arraystretch}{1.3}
    \caption{Summary of research works exploring LLMs for various dialogue tasks.}
    \label{tab:llm_ds}
    \centering
    \begin{adjustbox}{max width=0.8\textwidth}
    \begin{tabular}{c|c|c|c|c|c|c|c}
    \hline
    Method(s) & Cue(s)\dag & Type & Task(s)* & LLM(s) & Settings & Language & Domain \\
    \hline
    Pan et al. \citep{pan2023preliminary} & S & \multirow{4}{*}{TOD} & NLU & ChatGPT & ZS & EN & General \\
    Michael et al. \citep{heck-etal-2023-chatgpt} & S & & DST & ChatGPT & ZS & EN & General\\
    IT-LLM \citep{hudevcek2023large} & S & & DST, NLG & ChatGPT & ZS, FS & EN & General  \\
    SGP-TOD \citep{zhang2023sgptod} & S & & DST, DPL, NLG & ChatGPT & ZS & EN & General \\
    \hline
    GDP-ZERO \citep{yu2023promptbased} & S & & DPL & ChatGPT & FS & EN  & PersuasionForGood \\
    Qin et al. \citep{qin2023chatgpt} & S & \multirow{6}{*}{ODD} & DS & ChatGPT & ZS & EN & General \\
    ProCoT \citep{deng2023prompting} & S & & PD & ChatGPT, Vicuna & ZS, FS & EN & General \\
    Fan et al. \citep{fan2023uncovering} & S & & DDA & ChatGPT & ZS, FS & CN, EN  & General\\
    Huynh et al. \citep{huynh2023understanding} & S & & DE & GPT3, BLOOM & FS & EN  & General \\
    Mendonça et al. \citep{mendonça2023simple} & S & & DE & ChatGPT & ZS & Multilingual  & General  \\
    \hline
    Bang et al. \citep{bang2023multitask} & S & TOD, ODD & DRG & ChatGPT & ZS & EN  & General \\
    \hline
    CueCoT \citep{cuecot} & Per, Psy, E & \multirow{2}{*}{ODD} & DQA, DRG & ChatGPT, ChatGLM & ZS, FS & CN, EN & General  \\
    Zhao et al. \citep{zhao2023chatgpt} & E & & DU, DRG & ChatGPT & ZS & EN  & General \\
    \hline
    SMILE \citep{qiu2023smile} & Psy & ODD & DRG & \xmark &  \xmark & CN  & Mental Health \\
    \hline
    ExTES \citep{zheng2023building} & E & \multirow{4}{*}{ODD} & DRG & LLaMA & Finetuning & EN & General \\
    ChatCounselor \citep{liu2023chatcounselor} & Psy & & DRG & Vicuna & Finetuning & EN & Mental Health \\
    CharacterChat \cite{tu2023characterchat} & Per, M & & DRG & LLaMA & Finetuning & EN & General \\
    Fu et al. \citep{fu2023reasoning} & E & & CE, DRG & ChatGPT, T5 & FS, Finetuning & EN & General \\
    \hline
    \multicolumn{8}{l}{\dag S--Semantics, Per--Persona, Psy--Psychology, E--Emotion, M--Memory} \\
    \multicolumn{8}{l}{* \texttt{DS}--Dialogue Summarization, \texttt{NLU}--Natural Language Understanding, \texttt{DST}--Dialogue State Tracking, \texttt{PD}--Proactive Dialogues, \texttt{DDA}--Dialogue Discourse} \\
    \multicolumn{8}{l}{Analysis, \texttt{DE}--Dialogue Evaluation, \texttt{DRG}--Dialogue Response Generation, \texttt{DQA}--Dialogue Question Answering, \texttt{DU}--Dialogue Understanding, \texttt{CE}--Causality} \\
    \multicolumn{8}{l}{Explanation} \\
    \end{tabular}
    \end{adjustbox}
    \vspace{-4mm}
\end{table*}

One crucial aspect is equipping dialogue systems with human-like perception abilities and reasoning abilities~\citep{wang2025selfreasoninglanguagemodelsunfold}. These abilities play a key role in comprehending the internal status of the dialogue context, which stands for any information that can be reasoned and deduced from semantic context during the conversation, benefiting several downstream dialogue tasks: dialogue understanding \citep{zhao2023chatgpt, zhang2024transfertod}, dialogue state tracking \citep{hudevcek2023large,heck-etal-2023-chatgpt,zhang2023sgptod, zhang2025pfdial}, emotional and personalized dialogue generation \citep{cuecot,zheng2023building,fu2023reasoning}, dialogue summarization \citep{qin2023chatgpt} and so on. Regarding various prompting methods to reason different aspects of internal status exhibited in the dialogue context, we divide these methods into two categories: 1) \textit{semantic perception}, which directly extracts useful information from the dialogue context in the required format; 2) \textit{semantic reasoning}, which reason and deduce some underlying information beneath the dialogue context to better understand user intention. Table~\ref{tab:llm_ds} shows recent progress in this field.

\subsubsection{Semantic Perception}
There are some works that attempt to evaluate the performance of LLM at various dialogue understanding tasks such as NLU, DST, and dialogue discourse analysis tasks. For example, Pan et al. \citep{pan2023preliminary} and Michael et al. \citep{heck-etal-2023-chatgpt} evaluate the capability of \texttt{ChatGPT} to compress the dialogue context into the semantic frame (structured output of NLU) and belief states (structured output of DST) in the context of TOD, respectively. Furthermore, Fan et al. \citep{fan2023uncovering} investigate \texttt{ChatGPT}'s performance in three discourse analysis tasks: topic segmentation, discourse relation recognition, and discourse parsing, of varying difficulty levels. The key is to compress the dialogue context, the unstructured text into structured data, to meet the requirement of downstream applications or tasks. On the other hand, another line of work tries to compress the lengthy dialogue context while keeping important semantic information, particularly in dialogue summarization tasks \citep{qin2023chatgpt}.

\subsubsection{Semantic Reasoning}
Besides semantics information, it is widely acknowledged that dialogue contexts generally convey a lot of information about the user status from a linguistic perspective \cite{tausczik2010psychological}. Specifically, the linguistic cues underlying dialogue context have been shown to be an effective means of revealing the emotions \cite{ekman1971universals}, personality traits, psychological characteristics \cite{tausczik2010psychological}, and other relevant information of users \cite{turney2002thumbs}. Consequently, recognizing and understanding these cues exhibited in the context of dialogues becomes crucial to comprehend user intentions and status \cite{cuecot}. By doing so, a dialogue system can generate responses that align with the user's expectations, meet their unique needs and preferences, and ultimately strive toward constructing a human-like conversational experience \cite{cuecot,salemi2023lamp}. Each linguistic cue has garnered considerable scholarly attention, resulting in a wealth of research and a variety of methodological approaches inspired by them. Specifically, there are three major types of cues center around the user and there are some studies attempt to seamlessly integrates these multiple linguistic cues into a cohesive conversational framework.

\begin{itemize}
    \item \textit{Psychology.} In the realm of mental health and counseling, it is crucial to consider the psychological state of user and then tailor the response to provide mental support~\citep{psyqa}. Recently, Liu et al. (2023) introduced ChatCounselor, a model designed to provide mental health counseling. Meanwhile, Qiu et al. (2023) leveraged ChatGPT to transform a QA dataset into multi-turn mental health dialogues, enabling more effective fine-tuning on specialized datasets.

    \item \textit{Emotion.} Emotional intelligence has always been a crucial feature of dialogue systems, enabling them to understand and generate more empathetic and contextually appropriate responses~\citep{wang2023emotional}. For examples, Fu et al. \citep{fu2023reasoning} integrates commonsense-based causality explanation, considering both the user’s perspective
    (user’s desires and reactions) and the system’s
    perspective (system’s intentions and reactions). Zhao et al. \citep{zhao2023chatgpt} evaluate the performance of \texttt{ChatGPT} on emotional dialogue in terms of various understanding and generation tasks such as emotion recognition, emotion cause recognition, dialogue act classification (emotion dialogue understanding), empathetic response generation and emotion support generation. In order to navigate limited data and non-standardized training in emotional dialogues, the ExTensible Emotional Support dialogue dataset (ExTES) is curated by utilizing the in-context learning potential of \texttt{ChatGPT} that synthesizes human insights.  

    \item \textit{Personality.} The way one speaks reflects their personality. It is straightforward to infer the user persona from textual information especially the dialogue context~\citep{cuecot}. In detail, Tu et al. \citep{tu2023characterchat} present CharacterChat, an extensive conversational system designed to provide social support. CharacterChat is comprised of a conversational model that is guided by personas and memory, alongside an interpersonal matching plugin model. This plugin model is responsible for identifying and assigning the most suitable supporters from the MBTI-1024 Bank to individuals based on their specific personas. 

    \item \textit{Unification.} Instead of considering different cues in isolation, Wang et al. \citep{cuecot} seamlessly blend them all into one cohesive conversational flow, regarding different aspects of cues as part of user status and prompt the LLMs to reason user status exhibited in the dialogue context, aiming to generate more helpful and acceptable responses for users.
    
\end{itemize}

Besides different linguistic cues, it is noteworthy that users may occasionally fail to furnish adequate semantic information within the dialogue context, resulting in ambiguous queries or unreasonable requests. Both of these are considered key aspects of a conversational agent’s proactivity \citep{DBLP:conf/ijcai/0002LLC23, deng2023prompting}. To trigger the proactivity of LLMs,
ProCoT \citep{deng2023prompting} is proposed to augment LLMs with the goal planning capability over descriptive reasoning chains, specifically focusing on three
key aspects of proactive dialogues: clarification, target-guided, and non-collaborative dialogues.

\subsection{External Acting}

\label{external_llm_ds}
To ensure the delivery of up-to-date, informative, trustworthy, and personalized responses, it is necessary to learn and plan how to interact with external different knowledge sources \citep{wang2023large, wang2025actingreasoningmoreteaching}. Therefore, the integration of these knowledge sources not only enhances the conversational prowess of the LLM-based dialogue system but also enables them to remain adaptable and aligned with the dynamic nature of the information they handle, ultimately elevating their utility and reliability in real-world scenarios. In this section, we mainly present how LLM-based dialogue systems interact with external environment to integrate with different knowledge sources and plan their usage.

\begin{itemize}
    \item \textit{Database.} TOD mostly relies on one specific database such as the restaurant or flight database to answer the user query, providing the up-to-date and necessary information to accomplish the task \citep{hudevcek2023large, zhang2023sgptod}. Recently, Hudevceke et al. \citep{hudevcek2023large} and Zhang et al. \citep{zhang2023sgptod} consecutively evaluate the performance of \texttt{ChatGPT} on the E2E TOD. In detail, Hudevceke et al. \citep{hudevcek2023large} introduce a novel approach for dialogue state tracking, where belief states are derived from dialog history by utilizing slot descriptions as prompts, subsequently leading to database entry retrieval and response generation. In contrast, SGP-TOD \citep{zhang2023sgptod} employs slot names and value examples, rather than slot descriptions, as prompts to facilitate frozen LLMs in generating belief states, thereby reducing human effort. In addition, they offer a policy skeleton within the Policy Prompter to guide LLMs in producing responses that comply with the predefined dialog policy, resulting in better performance than IT-LLM \citep{hudevcek2023large}. In order to maximize the utilization of databases within dialogue systems, ChatDB \citep{hu2023chatdb} employs large-scale models to generate SQL instructions during the dialogue process. Moreover, recent studies try to formulate the API call as tool utilization problems such as AppBench~\citep{wang-etal-2024-appbench} and DilogTool~\citep{wang2025rethinkingstatefultooluse}.

    \item \textit{Memory.} The importance of memory knowledge sources in the context of large language models (LLMs) for open-domain dialogue systems is paramount. These memory sources enable LLMs to retain and access relevant information from previous interactions, facilitating more coherent and contextually aware responses in long-term conversations. Effective memory mechanisms are crucial for a wide range of applications, including personal companion systems, psychological counseling, and secretarial assistance, where sustained interactions and the ability to recall past information are vital for providing a rich and personalized user experience \citep{lee2023prompted, zhong2023memorybank,wang2023recursively}. Particularly, Zhong et al. \citep{zhong2023memorybank} introduce MemoryBank, a novel memory mechanism inspired by the Ebbinghaus Forgetting Curve theory. MemoryBank allows LLMs to summon and selectively update memories, adapting to a user's personality over time. It demonstrates the utility of MemoryBank in a long-term AI companion scenario. Wang et al. \citep{wang2023recursively} introduce a recursive memory generation approach, where LLMs are trained to memorize small dialogue contexts and generate new memory as conversations progress, aiming to improve consistency in long-context conversations. Furthermore, Lu et al. \citep{lu2023memochat} present MemoChat, a pipeline for refining instructions to enable LLMs to use self-composed memos to maintain consistency in long-range conversations. It focuses on the iterative ``memorization-retrieval-response" process and incorporates tailored instructions. These works collectively contribute to addressing the challenge of long-term memory in LLMs with distinct approaches and innovations.
    
    \item \textit{Internet.} One key notorious problem of LLMs is the hallucinations \citep{Huang_hallucination_survey}\footnote{Most works about the hallucination can be found in \S~\ref{discuss_hallucination}.}. Prior studies demonstrate that external document sources can be used to validate the factuality of generated response \citep{wang2023survey} and thus reduce the hallucinations \citep{shuster2021retrieval, Huang_hallucination_survey}. In addition, document sources can be used to store up-to-date information, enabling the LLM to sync the latest world knowledge such as \texttt{New Bing}. While responses of LLM-based DS are more likely to include information not present in the relevant document, possibly including the presence of hallucinations, they are preferred by human judges compared with other baselines without using LLM and even human responses \citep{DBLP:journals/corr/abs-2309-11838}. To leverage enhancing the quality and reliability of responses together, \texttt{WebGLM} \citep{liu2023webglm} extends LLMs with web search and retrieval capabilities, demonstrating the efficiency and effectiveness in question-answering tasks, while Semnani et al. \citep{semnani2023wikichat} mainly leverage documents from Wikipedia to enhance the factuality of LLM-based DS, focusing on providing accurate and verified information with additional fact-checking and refinement steps. Furthermore, there are some works that explore the relationship between multiple documents \citep{trivedi-etal-2023-interleaving} or queries \citep{wang2025selfdcreasonactself} to solve one complex question requiring multi-hop reasoning.

    
    \item \textit{Profile.} Salemi et al. \citep{salemi2023lamp} first introduce the LaMP benchmark — a novel benchmark for training and evaluating LLMs for producing personalized outputs. LaMP offers a comprehensive evaluation framework with diverse language tasks and multiple entries for each user profile, consisting of three personalized understanding and four personalized generation tasks.
    
    \item \textit{Others.} Besides these common knowledge sources, there are other knowledge sources in a different organization or structure that also play a key role during response generation such as API \citep{li2023apibank,patil2023gorilla,liang2023taskmatrixai}, and Knowledge Graph \citep{bang2023multitask}.

    \item \textit{Unification.} Most existing LLM-based DS either focus on a single knowledge source or overlook the relationship between multiple sources of knowledge, which may result in generating inconsistent or even paradoxical responses \citep{majumder-etal-2020-like,fu-etal-2022-thousand}. To incorporate multiple sources of knowledge and relationships between them, Wang et al. \citep{wang2023large} propose SAFARI, a novel framework that takes advantage of the exceptional capabilities of large language models (LLM) in planning, understanding, and incorporating in both supervised and unsupervised settings. Specifically, SAFARI decouples the knowledge grounding into multiple knowledge sources selection and response generation, which allows easy extension to various knowledge sources including the possibility of not using any sources. 
\end{itemize}

\subsection{Internal Reasoning with External Acting}
\label{internal_exteal_llm_ds}

By focusing on enhancing the cognitive understanding of the dialogue context and allowing seamless interactions with external sources, LLM-based systems can become more sophisticated, effective, and capable of delivering personalized, trustworthy, helpful, up-to-date responses to various user queries and requests \citep{wang2023interactive, acikgoz2025desideratumconversationalagentscapabilities}. For examples, Tu et al. \citep{tu2023characterchat} utilize persona (internal) and profile memory (external) simultaneously to generate personalized responses while Xu et al. \cite{dulemon} alternatively take advantage of short-term memory (i.e., internal reasoning) and long-term memory (i.e., external source). Within the multi-persona collaboration framework TPE (Thinker, Planner, and Executor), Wang et al. \citep{wang2023tpe} leverage the internal status of dialogue context to guide the planning use of different external knowledge sources, regarding different sources as one specific type of conceptual tools. In addition, Sumers et al. \citep{sumers2023cognitive} propose \texttt{CoALA}, which delineates a language agent that incorporates modular memory components, a structured action space for interactions with both internal memory and external environments, and a generalized decision-making process for action selection.

\begin{table*}[!t]
    \renewcommand{\arraystretch}{1.0}
    \caption{Performance of LLM-based Dialogue System which blender internal reasoning and external acting in an unified framework. The results are copied from CoALM ~\citep{acikgoz2025singlemodelmastermultiturn}.}
    \label{tab:colam_results}
    \centering
    \begin{adjustbox}{max width=0.6\textwidth}
    \begin{tabular}{ll|cc|cc}
    \toprule
    \multirow{2}{*}{\textbf{Type}} & \multirow{2}{*}{\textbf{Method}} & \multicolumn{2}{c|}{\textbf{MultiWoZ 2.4}} & \multicolumn{2}{c}{\textbf{API-Bank}} \\

    \cline{3-6} & & Success & JGA & Rouge-L & BLEU-4 \\
    
    
    \hline
    \multirow{2}{*}{Trad. DS} &
    FNC-ToD & 44.4 & 37.9 & 3.9 & 3.3 \\
    & NC-Latent-ToD & 68.3 & 39.7 & 3.2 & 0.8 \\
    
    \hline
    \multirow{3}{*}{LLM-Prompting} & 
    Llama 3.1 8B Instruct & 19.9 & 26.3 & 72.7 & 62.3 \\
    & Granite-20B-Code & 10.7 & 21.8 & 60.3 & 43.8 \\
    & GPT4o-mini & \textbf{69.9} & 38.4 & - & - \\

    \hline
    \multirow{4}{*}{LLM-Finetuning} &
    Hammer 2.0 7B & 23.5 & 21.7 & 90.1 & 85.4 \\
    & ToolAce & 18.0 & 34.4 & 81.5 & 76.1 \\
    & \cellcolor{purple!10}CoALM (8B) & \cellcolor{purple!10}51.6 & \cellcolor{purple!10}30.4 & \cellcolor{purple!10}92.8 & \cellcolor{purple!10}89.4 \\
    & \cellcolor{purple!10}CoALM (70B) & \cellcolor{purple!10}69.4 & \cellcolor{purple!10}\textbf{43.8} & \cellcolor{purple!10}\textbf{92.7} & \cellcolor{purple!10}\textbf{89.5} \\

    \bottomrule
    \end{tabular}
    \end{adjustbox}
\end{table*}

Despite these encouraging advancement to bridge the internal reasoning and external interaction, the better unification remains an open challenge due to the complex relation between them. Sometimes their functionalities may overlap with each other when one problem can be solve by both internal reasoning or external acting. Wang et al.~\citep{wang2025selfdcreasonactself} propose Self-DC which unify reasoning and acting as different functions and dynamically invoke them based on self-aware knowledge boundary of LLMs while SMART~\citep{qian2025smartselfawareagenttool} further encompass a wider range of tool utilization scenarios, with a particular focus on addressing the issue of tool overuse for existing language agents. Besides that, there also exist conflict knowledge or results between internal reasoning and external acting~\citep{xu-etal-2024-knowledge-conflicts}, i.e., the conflict between the parametric knowledge and retrieved knowledge. More recently, Emre et al. \cite{acikgoz2024convagents} propose a concept of conversational agents, where an LLM-based dialogue system is designed to perform multi-turn interactions with users, by integrating reasoning and planning capabilities with action execution. Combining the advantages of both language agent and dialogue system, conversational agent is capable to master both multi-turn conversation and tool use with an unified model~\citep{acikgoz2025singlemodelmastermultiturn}. Table~\ref{tab:colam_results} shows the results between CoALM with other strong baselines. It can be drawn that CoALM achieves
leading performance on both TOD and LA benchmarks, revealing the strong robustness and generalization. In conclusion, ensuring seamless integration between internal reasoning and external acting enables better monitoring and controllability~\citep{wang2025theoryagentstoolusedecisionmakers}. Addressing these issues is essential for developing more reliable, adaptive, and intelligent LLM-based dialogue systems capable of engaging in complex and dynamic conversations.

\subsection{Summary}

Throughout the evolution of dialogue systems, we have consistently observed a clear trend: the pursuit of a powerful, scalable, and unified system. From the independent development of TOD and ODD to their gradual fusion at various levels, and now the rise of LLM-based dialogue systems, the ultimate goal remains the same. As a consequence, existing LLM-based dialogue systems enables more natural, personalized, and trustworthy interactions that adapt to complex real-world scenarios, as exemplified by architectures like TPE \citep{wang2023tpe} and CoALA \citep{sumers2023cognitive}. However, significant challenges persist. First, ethical concerns emerge regarding privacy preservation when handling sensitive user profiles (\S~\ref{external_llm_ds}) and psychological manipulation risks in emotionally-aware systems (\S~\ref{internal_llm_ds}). Second, one notable drawback is the issue of hallucinations, where the system may generate inaccurate or misleading information. Despite external knowledge source may alleviate the hallucinations in the response, it introduces additional hallucination issues in the planning step of dialogue system. Overall, while LLM-based dialogue systems have pushed the boundaries of what traditional dialogue system can achieve, further research is essential to ensure their reliability, ethical compliance, and practical deployment in complex environments.


\section{Discussions}
\label{discussion}

\subsection{Data}

As large language model (LLM)-based dialogue systems continue to evolve towards becoming integral components of everyday life and potential gateways to artificial general intelligence (AGI), the discussion around data becomes increasingly pivotal. Two central aspects of data issues emerge: general challenges related to data schema and quality, and specific challenges concerning diversity in terms of languages, domains and applications.

\subsubsection{General Data Issues: Universal Data Schema and Quality}

One key observation we can drawn from previous evolution of dialogue system is the unification of data schema or format is the very foundation which enables the the unification of tasks and models. As the LLM-based dialogue system become increasingly embedded in real-world applications to solve more and more complex task, it proposes new challenges about new data schema and format to learn and complete these new tasks together. On the schema side, several pioneer attempt to unify all tasks into one unified data format. For example, CoALM~\citep{acikgoz2025singlemodelmastermultiturn} adopts ReAct-style data schema to empower the LLM-based dialogue system capable to function as both dialogue system and language agent, aiming to unify multi-turn conversation, function calling and task completion. On the quality side, it is difficult and expensive to manually collect the required high-quality data for each task and application, and thus most of researcher try to utilize LLMs themselves to synthesize data~\citep{wang2023self}, leading in self-evolving LLMs~\citep{yuan2024selfrewardinglanguagemodels, tao2024survey}. For examples, Yuan et al.~\citep{yuan2024selfrewardinglanguagemodels} study Self-Rewarding Language Models, where the language model itself is used via LLM-as-a-Judge prompting to provide its own rewards during training, achieving iterative improvement by finetuning using self-rewarding self-generated responses, followed by lots of works~\citep{huang2025selfevolvedrewardlearningllms, hongruself}. Fruthermore, as the model approaches or surpasses human-level intelligence, its performance may become limited by the quality and availability of human-labeled data. This creates a challenge in scaling data acquisition, such as weak-to-strong supervision~\citep{burns2023weak}, making it an open question that remains to be addressed.

\subsubsection{Specific Data Issues: Diversity in Language, Application, and Domain} While the above general issues affect the performance of LLM-based dialogue systems, specific challenges arise when addressing diverse, low-resource languages and various application domains. Although we previously discussed existing work on low-resource languages, a significant gap remains in resources and applications compared to high-resource languages. This imbalance has far-reaching consequences, as dialogue systems may struggle to deliver equitable and culturally aware services, ultimately reinforcing existing inequalities and cultural biases. Addressing these challenges involves collecting and curating data that reflects the linguistic, cultural, and domain-specific diversity of real-world interactions. Additionally, dialogue systems designed for niche domains -- such as medical advice, legal consultations, or finance -- demand domain-specific data that accurately captures the specialized vocabulary, contextual nuances, and ethical considerations inherent to those fields. By emphasizing scalability and inclusivity in data collection and processing, dialogue systems can become more resilient, equitable, and better equipped to meet the diverse and complex needs of our interconnected world.

\subsection{Applications}

This section examines real-world applications of dialogue systems, categorized into three types based on use cases: \textit{Information-Seeking and Decision Support}, \textit{Task Orchestration and Execution}, and \textit{Affective and Recreational Engagement}. These systems are deployed using either standalone architectures, which operate independently, or integrated architectures, which connect to external tools, knowledge bases, or other components. We classify each system accordingly and evaluate their relative advantages. A key trend across all categories is the growing effectiveness of LLM-based dialogue systems when integrated with external resources. Recent systems like GPT-4~\cite{gpt4}, Claude~\footnote{\url{https://www.anthropic.com/news/claude-3-7-sonnet}}, and DeepSeek~\cite{deepseekai2025deepseekr1incentivizingreasoningcapability} exemplify this, functioning as general-purpose agents enhanced by access to specialized tools. Despite the rise of such versatile systems, specialized dialogue applications remain vital for scenarios with constraints such as limited compute, domain-specific needs, or strict privacy requirements.

\subsubsection{Information-seeking and Decision-support.}
This type of applications help users access information and make informed decisions through conversational interfaces. 
We introduce three representative applications: 1) \textit{E-commerce Customer Support.}
Dating back to the 1990s, \citet{meng1996wheels} introduced \textsc{WHEELS}, an \textit{Integrated} dialogue system using parsing strategy for NLU and form-filling for dialogue management, which functioned as a natural language interface to an automobile advertisement database. With the rise of deep learning, leading e-commerce companies adopted \textit{Standalone} models trained on large-scale corpus~\citep{chen2020jddc}. Latest implementations \textit{integrate} product knowledge bases and multimodal interactions, enhancing customer experience through personalized services while maintaining conversational coherence~\cite{gupta2024artificial}; 2) \textit{E-government Civil Service.} Recently, government agencies worldwide have widely adopted dialogue systems to help citizens access information and navigate services~\cite{watch2022national, abed2024understanding}. For instance, Singapore's GovTech team presents ``Ask Jamie''\footnote{\url{https://www.tech.gov.sg/media/technews/govtech-team-behind-ask-jamie-government-chatbot/}}, a chatbot to address citizen inquires for government agencies. Initially deployed as a \textit{standalone} system trained on extensive QA documents, latest government implementations now leverage \textit{integrated} approaches using RAG techniques with capable dialogue systems as the backbone\footnote{\url{https://insidegovuk.blog.gov.uk/2024/01/18/the-findings-of-our-first-generative-ai-experiment-gov-uk-chat/}}, effectively addressing data privacy concerns and mitigating hallucination issues; 3) \textit{Financial Robo-advisors}. Adopting dialogue systems in finance to provide asset and wealth management advice has gained significant traction with the advent of LLM-based solutions \citep{polireddi2024artificial}. In this sector, \textit{Integrated} systems that combine LLMs with domain-specific techniques to deliver sophisticated advisory services are prevalent. Leading companies like Wealthfront have incorporated dialogue systems into their automated investment platforms\footnote{\url{https://research.wealthfront.com/whitepapers/investment-methodology/}}, leveraging seamless interactive capabilities to deliver personalized financial guidance and optimize portfolio management at substantially lower costs than traditional human advisors. Beyond the previously mentioned fields, dialogue systems are also widely adopted in domains requiring specialized expertise including: medical advice systems (e.g., Google's Med-PaLM \citep{singhal2025toward}), educational assistants (e.g., Khanmigo\footnote{\url{https://www.khanmigo.ai/}}), and legal question-answering services \citep{yao2024lawyer}.



\subsubsection{Task Orchestration and Execution}

Dialogue systems is widely adopted to help users accomplish specific goals through conversational interfaces, ranging from simple \textit{standalone} helpers to complex \textit{integrated} systems that coordinate multiple tools and services. We review such applications in following two categories: 1) \textit{Personal Assistance}. 
Dialogue systems functioning as personal assistants are usually embedded in diverse smart devices, requiring tight integration between natural language processing capabilities and hardware interfaces or APIs. 
Early implementations 
like \texttt{Siri} and \texttt{Alexa} provides relatively basic command execution, while latest LLM-based systems such as Claude and ChatGPT  leverage more sophisticated semantic understanding and reasoning to deliver sophisticated assistance through advanced tool integration for personal task management; 2) \textit{Specialized Expert Assistance.}
Different from aforementioned personal assistants, such expert assistance DS needs to execute professional tasks with domain expertise.
For example, \texttt{DoNotPay}~\footnote{\url{https://donotpay.com/}} functions as a \textit{legal} assistant that generates properly formatted legal documents and navigates procedural requirements through conversational interfaces.
BloombergGPT~\cite{wu2023bloomberggpt} operates as a pre-trained \textit{standalone} LLM optimized for finance-specific tasks like news sentiment analysis.
FinRobot~\cite{yang2024finrobot} implements an \textit{integrated} architecture where the LLM serves as the ``brain'', i.e., a central orchestration module, for a financial agent, coordinating specialized tools for financial analysis, market prediction, etc. More recently, LLM-based DS have unlocked new frontiers in various fields, enabling autonomous problem-solving and complex decision-making in real-world tasks, such as ChatDEV~\cite{chatdev} for \textit{software development}, Copilot\footnote{\url{https://copilot.microsoft.com/}} for \textit{coding assistance}, and GUI agents~\cite{wang2024gui} that execute \textit{interface actions} based on user instructions.


\subsubsection{Affective and Recreational Engagement}

Dialogue systems for affective engagement serve dual purposes: facilitating recreational social interactions and providing emotional support through therapeutic conversations. 
Thus they operate on a spectrum from casual companionship to clinical intervention.
Commercial implementations like \texttt{Replika}~\footnote{\url{https://replika.com/}} primarily function as \textit{Standalone} systems offering \textit{companionship} through social chit-chat, maintaining substantial user engagement through simulated emotional bonds.
They also show clinical benefits, with research confirming their effectiveness in reducing suicidal thoughts through timely emotional support and crisis intervention. More specialized \textit{therapeutic} systems usually \textit{integrate} external tools to provide more professional service~\cite{filienko2024toward}. \texttt{WoeBot Health}~\footnote{\url{https://woebothealth.com/}} utilizes persistent databases to track user emotional states and deliver personalized mental health interventions, while \texttt{Wysa}~\footnote{\url{https://www.wysa.com/}} employs a hybrid approach connecting automatic conversations with emergency intervention APIs and human therapist escalation when needed. 


\subsection{Multi-modality}

In the ever-evolving landscape of conversational AI, the integration of multimodal capabilities has emerged as a pivotal advancement that promises to reshape the future of LLM-based dialogue systems \citep{liu2023visual,minigpt5}. Before the advent of large language models, dialogue systems primarily operated within the confines of text-based interactions, with few works explore the multimodal dialogue systems \citep{johnston-etal-2002-match, 9948337,electronics11203409,10.1145/3240508.3240605}. However, with the introduction of powerful models such as GPT-4 \citep{gpt4} and their ability to comprehend and generate text and even image at a human-like level, the stage was set for a transformative convergence of text, images, audio, and video in the realm of conversation \citep{liu2023visual}. In this section, we first briefly review the previous work before the era of large langauge models and then explore the exciting prospects and future directions of multimodal LLM-based dialogue systems, shedding light on their evolution, potential applications, and the dynamic impact they promise to have on a wide range of fields, from customer service and medicine and beyond.

\subsubsection{Before the era of LLMs}

Before the era of large language models, there were relatively few efforts which try to fuse different modalities seamlessly in the conversational flow. According to different modalities of input and output in dialogue systems, these works can be divided into two categories\footnote{\url{https://github.com/ImKeTT/Awesome-Multi-Modal-Dialog}}: 1) systems with diverse modalities in the input but not in the output \citep{das2017visual,alamri2019audiovisual,shuster-etal-2020-image,wang2021openvidial,zheng-etal-2022-mmchat}; 2) systems with multiple modalities in the output \citep{10.5555/3504035.3504121,zang-etal-2021-photochat,ma-etal-2022-unitranser,feng-etal-2023-mmdialog} \footnote{This includes systems with the incorporation of multiple modalities in both the input and output.}. Predominantly, a significant portion of research endeavors align with the first category, where input comprises text and image modalities, resulting in textual responses pertaining to the image content. A noteworthy milestone in this progression was achieved by Alamri et al. \citep{alamri2019audiovisual} when they pioneered the integration of audio information into the input phase, marking a notable enhancement in the multimodal dialogue system's capabilities. It is essential to highlight that the majority of efforts falling under the second category primarily involve the generation of output in alternative modalities, such as images, employing a discriminative approach by selecting one from existing sources rather than generating one \citep{johnston-etal-2002-match,10.1145/3240508.3240605}. Furthermore, some works explore multilingual situations in the context of multimodal dialogue \citep{hough-etal-2016-duel}.

\subsubsection{After the era of LLMs}

Since the emergence of LLMs, a notable trend has been the alignment of feature spaces across different modalities within dialogue systems. This alignment effort has taken two distinct directions: 1) the alignment of non-textual modalities with the textual feature space \citep{liu2023visual,maaz2023videochatgpt}; and 2) approaches that pursue alignment in the opposite direction \citep{girdhar2023imagebind}. These endeavors mark a significant shift in the field, aiming to harmonize the representation and understanding of diverse modalities within the context of LLM-based dialogue systems. In detail, LLaVA represents the inaugural endeavor in visual instruction tuning using data generated by language-only GPT-4, serving as an end-to-end trained large multimodal model that connects a image encoder and LLM for general-purpose visual and language understanding \citep{liu2023visual}. Despite some preliminary efforts have been made in the development of image-based dialogue systems, some works specifically delve into the relatively uncharted territory of video-based dialogue systems \cite{maaz2023videochatgpt}. For example, Muhammad et al. \citep{maaz2023videochatgpt} introduce Video-ChatGPT, which is a multimodal model that merges a video-adapted visual encoder with a LLM by simply finetuning a linear layer.

In closing, it is crucial to acknowledge that while the evolution of multimodal dialogue systems has been marked by remarkable progress, there still are many challenges and problems. As we forge ahead, we confront an array of practical obstacles that necessitate rigorous research and innovation. The alignment of feature spaces between modalities continues to be a substantial concern, as ensuring a seamless interaction between various data types remains an ongoing challenge. Furthermore, the interpretability and explainability of these complex multimodal systems pose questions regarding their reliability and ethical use. These challenges underscore the need for continued exploration, experimentation, and the development of novel approaches to make multimodal dialogue systems not only powerful but also robust, accessible, and ethically sound, thus solidifying their place in the future of AI-driven conversational technology.

\subsection{Hallucination}
\label{discuss_hallucination}

Language model-based dialogue systems have demonstrated significant potential across a broad range of applications. However, an increasing concern is their observed inclination towards hallucinations \citep{Huang_hallucination_survey, ji2023survey}. These instances, where system-generated responses defy universally accepted word knowledge, present a substantial challenge to the reliability and effectiveness of these systems in real-world scenarios. As the utilization of these dialogue systems continues to evolve, understanding and mitigating the impact of these hallucinations becomes a critical task for their successful deployment.

\subsubsection{Understanding the Causes of Hallucinations}

Understanding the underlying causes of hallucinations in language model-based dialogue systems is a crucial first step toward devising effective mitigation strategies. The hallucination phenomenon often results from a complex interplay of several factors. Inadequate training data \citep{kandpal2023large, xue2023improving} can play a significant role—when the training data lacks diversity or doesn't cover certain topics, the model might attempt to fill in the gaps in its knowledge, leading to hallucinated responses. Model overgeneralization \citep{hupkes2020compositionality, lin2022truthfulqa} is another major contributor, where an language model might overgeneralize from the patterns it has learned during training, leading to responses that may seem plausible but are actually incorrect or nonsensical in a given context. Lastly, the inherent randomness in the generation process can also lead to hallucinations. The stochastic nature of language models \citep{ranzato2015sequence, lee2023factuality} means they can sometimes generate incorrect text, particularly when the model is unsure about the most appropriate response. Understanding these underlying causes provides a foundation to explore potential mitigation strategies.

\subsubsection{Detecting Hallucinations in Dialogue System}

Research on hallucination detection has evolved along four main trajectories. Firstly, human evaluations have been conducted to assess the model output from diverse perspectives \citep{yu2023generate,liu2023evaluating}. However, these evaluations, while valuable, encounter issues due to their time-consuming nature and the inherent subjectivity of human judgement, leading to scalability and reproducibility challenges. 
Secondly, datasets have been constructed to evaluate the actuality of open-domain generation with the aid of references \citep{lee2023factuality}. These methods, while offering a more objective evaluation approach, are limited by their dependence on human-labelled references, which can negatively impact their applicability in real-world scenarios and their generalizability to dynamically generated content. 
Thirdly, self-evaluation methods \citep{varshney2023stitch} have been developed to estimate a model's uncertainty in its output. Despite the simplicity and appeal of these methods, they often lack interpretability and are less effective when dealing with long-form responses. 
Fourth, some recent studies \citep{min2023factscore} have applied fact-checking principles to spot factual inaccuracies in the model's output. These efforts, however, are often constrained by the limitations of external knowledge bases used in the fact-checking process.
Finally, an emergent line of research focuses on integrating references or fingerprints into the generated content \citep{gao2023enabling}. This approach, while promising, introduces a layer of traceability and verifiability to the output, enabling users or regulators to trace and verify the origins and accuracy of the content.

\subsubsection{Strategies for Mitigating Hallucinations}

To mitigate hallucinations in language model-based dialogue systems, researchers have proposed several methods, which can be broadly categorized into five main approaches. 
One prominent approach is retrieval-augmentation \citep{shuster2021retrieval}, which involves incorporating information from external sources during the generation process to ensure the accuracy of the output.
Model editing is another approach adopted by researchers to mitigate hallucinations in language model-based dialogue systems. This involves precisely adjusting the model's parameters to correct erroneous knowledge inherent in the system.
Post-editing \citep{dziri-etal-2021-neural} is another commonly used approach to mitigate hallucinations in dialogue systems. After the model's output is generated, post-editing techniques are applied to correct any detected hallucinations. This typically involves rule-based or pattern-based modifications to the generated output, ensuring that the final response is both coherent and accurate. 
Indeed, some methods aim to mitigate hallucinations by reducing the randomness in language model decoding. For instance, dynamic top-p sampling decoding \citep{lee2023factuality} dynamically adjusts the randomness as the decoding process progresses, thus reducing the element of chance and helping to decrease the incidence of hallucinations.
Lastly, better training strategies, such as RLHF \citep{xue2023improving}, have been developed to encourage the model to avoid making factually incorrect statements. By leveraging feedback from humans during the training process, these methods can guide the model towards generating more accurate and reliable responses.

Despite considerable progress in the analysis, detection, and mitigation of hallucinations, they remain a frequent issue in language model-based dialogue systems. Several specific challenges persist. For instance, the wide range of hallucinations, from minor inaccuracies to major factual errors, complicates detection and correction. Furthermore, the lack of large, high-quality datasets for training and evaluation hampers the development of more effective solutions. Balancing creativity and factual consistency is another challenging aspect, as overcorrection could lead to less engaging and overly cautious responses. Lastly, the need for real-time responses in many applications poses a significant challenge, as comprehensive hallucination detection and mitigation can be computationally expensive. Given these challenges, ongoing research and innovation are crucial to improve the effectiveness of these systems in practical applications.

\subsection{Safety}
Alongside the accelerated evolution of language models and dialogue systems, emerging safety concerns become increasingly crucial.
Given the indispensable social impact of current LLM-based dialogue systems, their safety issues have garnered attention not only from the NLP research community~\cite{deng2023recent, bai2022constitutional} but also from industry, philosophers, social scientists~\cite{irving2019ai}, policymakers~\cite{zeng2018linking}, etc.
In this subsection, we begin by introducing the safety problem and its distinctive nature in dialogue systems which are directly interacting with human users.
Subsequently, we offer a brief overview of existing safeguarding methods and discuss their limitations.

\subsubsection{Understanding the Safety Issue}
The term \textit{safety} encompasses a broad spectrum of issues~\citep{weidinger2022taxonomy}, comprising, but not limited to, toxic language~\cite{deng2022cold}, social bias~\citep{sap2020social,zhou-etal-2022-towards-identifying}, problematic moral or values inclination~\citep{kim2022prosocialdialog}, and misleading or risky suggestions~\cite{mei2022mitigating,sun2022safety}.
It is noteworthy that the rigorous scope and taxonomy are still an ongoing discussion. Moreover, it is anticipatable that as the system's capability and application scenarios broaden, the community may encounter novel safety challenges.
For example, there is an emergent research trend of investigating prosocial dialogue systems~\cite{kim2022prosocialdialog}, which are sensitive to social factors such as social norms~\cite{kim2022soda}, moral stances~\cite{sun2023moral}, and human values~\cite{hendrycks2023aligning}. 
This progress stems from the dialogue systems' expanding capabilities to engage in democratic conversations that involve personal feelings, desires, and social relationships.
In light of this, it is crucial to consistently evaluate the performance of dialogue systems to swiftly detect and tackle any new safety concerns that may arise.

\subsubsection{Existing Safeguarding Methods}
Learning from enormous corpora from various sources, current dialogue systems inevitably learn undesired behavior.
Mainstream safeguarding techniques can be categorized into two types: \textit{baked-in safety}~\cite{xu2020recipes} and \textit{pipeline safety}~\cite{deng2023recent}.
Baked-in safety refers to the methods that change the model at the training or decoding stage~\cite{schick2021self}.
The majority of open-source LLMs have incorporated various safety training methods during the training or fine-tuning stages~\cite{thoppilan2022lamda, bai2022constitutional}, with the RLHF technique~\cite{instructgpt} being the most effective and widely-discussed option.
This method can embed safety into models, however, it is not flexible nor explainable.
Pipeline safety does not require modification to models, instead, it uses an add-on safety layer (usually a safety detector) to block unsafe input or output~\cite{dinan2022safetykit}.
Nevertheless, this method cannot defend against the malicious use of open-sourced models. Another common and important obstacle in the above methods is the requirement of well-annotated corpora to train classifiers. Recently, there has been a novel research line utilizing the instruction following the ability of LLMs to perform self-critique~\cite{wang2023selfcritique} and self-correct~\cite{ganguli2023capacity}.
The LLMs are instructed to serve as the classifier and make post-edition to its responses based on the classification results. Moreover, it is also instructed to self-redteaming~\cite{wang2023selfguard} to explore high inductive user inputs and enhance the robustness automatically.

\subsubsection{Challenges and Discussions}
The trade-off between safety and system performance has long been an intricate issue during the development of safe dialogue systems. 
\citep{xu2020recipes} highlights the challenges of simultaneously maintaining safety and engagement in a dialogue system. 
Adding to this, Bai et al. \citep{bai2022training} provides evidence that there exists a tension between ensuring harmlessness, i.e., safety, and maintaining helpfulness in such systems. 
This tension, further referred to by researchers as the \textbf{\textit{alignment tax}}~\cite{askell2021general} remains an unresolved issue in the field. Another challenge is that the safety standards can change over time, culture, etc.~\cite{sun2023moral}
For example, \textit{mimic human} was an important objective for old-days chatbots~\cite{zhou2020design}.
However, nowadays systems are trained to avoid anthropomorphism~\cite{glaese2022improving}. 
Furthermore, the criticism of generalizing the stances and values of a limited group of annotators has not been settled.
Considering this, the methods that require a huge amount of training data and extensive training procedure is inflexible.
A growing body of literature~\cite{zhou2023rethinking} advocates for steering clear of aligning AI models with specific values, since the human influence during the development lifecycle can inadvertently introduce systematic biases~\cite{sap2021annotators, fraser2022does}.
Instead, scholars argue that shaping AI output through transparent and adaptable guidelines would be a more prudent approach~\cite{zhou2023rethinking}. 
However, the exploration of this direction remains in its infancy and warrants further investigation to develop a safe dialogue system.

\section{Conclusion}
\label{conclusion}

This survey provides a comprehensive overview of the evolution of dialogue systems alongside advances in language modeling-from Statistical Language Models (SLMs) to Neural Language Models (NLMs), Pre-trained Language Models (PLMs), and the current generation of Large Language Models (LLMs). We begin by outlining the contextual background of dialogue systems, defining four developmental stages to trace the progression of dialogue systems, highlighting clear bottom-up evolution and clear path for future direction. Our discourse extends to the challenges and applications of LLM-based dialogue system in real-world tasks.

\bibliographystyle{ACM-Reference-Format}
\bibliography{custom}

\appendix
\clearpage



\end{document}